\documentclass[11pt]{article}

\usepackage[preprint]{acl}

\usepackage{times}
\usepackage{latexsym}
\usepackage{booktabs}
\usepackage{multirow}
\usepackage{amsmath}
\usepackage{tcolorbox}
\usepackage[T1]{fontenc}

\usepackage[utf8]{inputenc}

\usepackage{microtype}

\usepackage{inconsolata}

\usepackage{graphicx}

\def\eg{{\it e.g., }}

\title{Cross-Centaur: Modeling Individual Decision-Making Across Psychological Tasks with Language Models}

\author{
  \textbf{Hikaru Asano\textsuperscript{1}}\thanks{Equal contribution. Details in Appendix~\ref{sec:contribution}},
  \textbf{Hidezo Suganuma\textsuperscript{1}}\footnotemark[1],
  \textbf{Ryutaro Mori\textsuperscript{2}}\footnotemark[1],
  \textbf{Yukino Baba\textsuperscript{1}}
\\
\\
  \textsuperscript{1}The University of Tokyo,
  \textsuperscript{2}RIKEN Center for Brain Science
\\
  \small{\texttt{\{asano-hikaru19,suga-zo0528utokyo,yukino-baba\}@g.ecc.u-tokyo.ac.jp}}\\
  \small{\texttt{ryutau.mori@gmail.com}}
}

\begin{document}
\maketitle
\begin{abstract}
    Large language models (LLMs) are increasingly used as computational models of human behavior. Recent models such as Centaur have advanced this goal by fine-tuning LLMs on trial-by-trial decisions from psychological experiments. However, existing large-scale behavioral datasets do not link the same participants across different tasks, so these models learn population-level regularities but are not trained to model individual-level regularities across tasks. We introduce Cross-Psych-101, a dataset of 591 participants and 261{,}551 valid decisions across 12 Bandit, Tradeoff, Structure, and Social tasks. We train Cross-Centaur to predict a participant's target-task actions from a target-task description and the source-task description paired with the same participant's source-task behavior. Cross-Centaur outperforms off-the-shelf LLMs and Centaur on held-out participants, both within and across tasks. Probing analyses suggest that its hidden states encode participant-level behavioral tendencies that transfer across tasks. These results highlight the potential of cross-task individual-level data for predicting human decision-making with language models.\footnote{Cross-Pysch-101 is available at \url{https://huggingface.co/datasets/HikaruAsano/cross-psych-101}.}
\end{abstract}

\section{Introduction}

\begin{figure}[t!]
    \centering
    \includegraphics[width=\columnwidth]{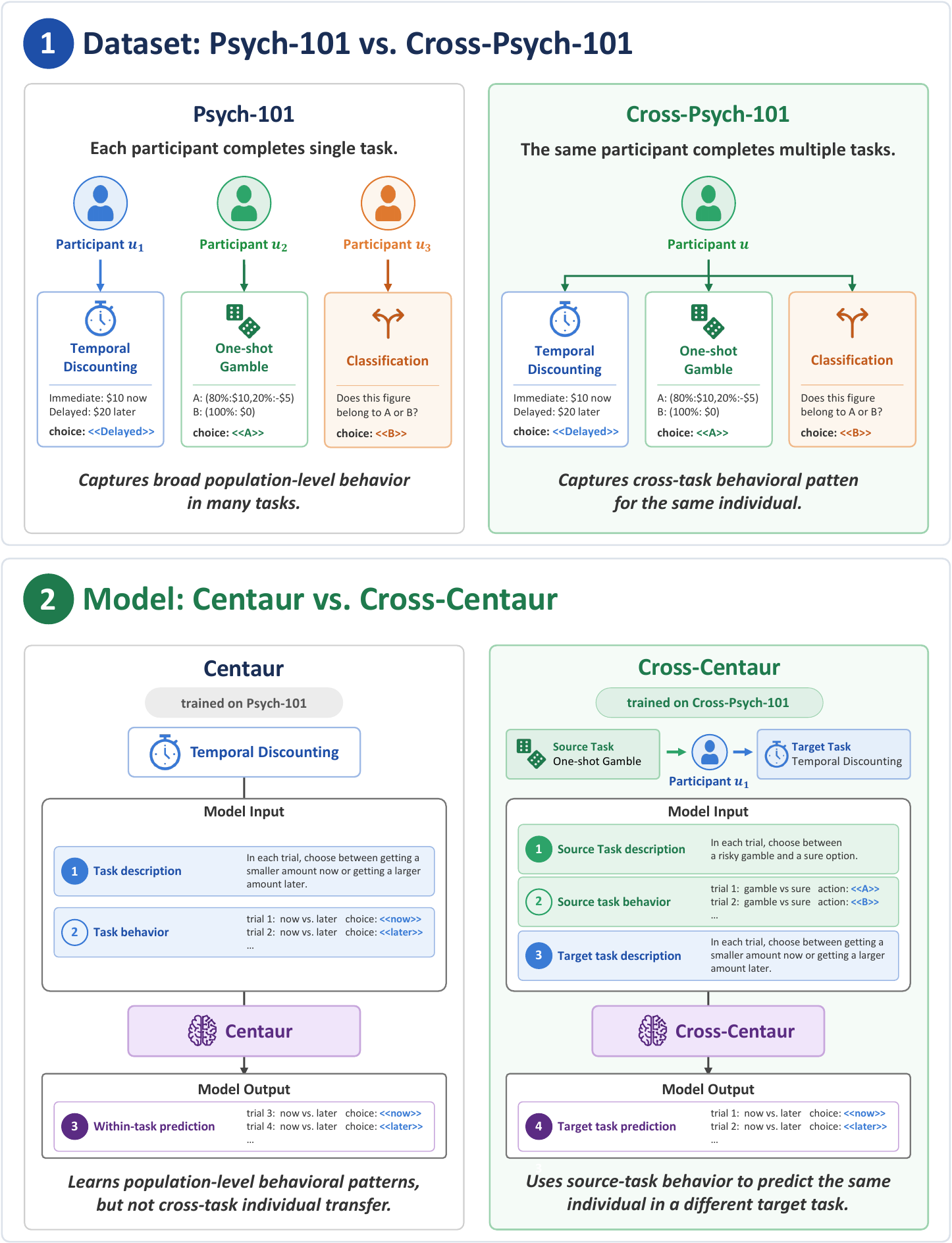}
    \caption{
    Comparison of Psych-101 and Centaur with Cross-Psych-101 and Cross-Centaur. Psych-101 observes participants within individual tasks and supports within-task behavioral prediction, whereas Cross-Psych-101 follows the same participants across tasks, enabling Cross-Centaur to use source-task behavior to predict the same individual's behavior in a different target task.
    }
    \label{fig:teaser}
\end{figure}

Large language models (LLMs) are increasingly used to model human decision-making and build realistic multi-agent simulations~\citep{Aher2023-nr,Filippas2024-lx,Park2023-ke,Gao2024-uc,Zhou2025-qa}. Because many tasks and responses can be described in natural language, LLMs provide a common interface for studying behavior across domains such as learning, decisions under uncertainty, and economic games~\citep{Demszky2023-ez,Ziems2024-sa,Binz2024-vx,Binz2025-jo}.

However, off-the-shelf LLMs are trained mainly to generate text and follow instructions, not to predict or replicate human decision-making~\citep{Ouyang2022-gr}.
Prior work has accordingly documented various behavioral divergences between LLMs and humans in decision-making and cognitive biases~\citep{cheung2025amplified,Liu2025-aa,Echterhoff2024-gm}.
LLM-based behavioral models should thus be grounded in empirical human data and evaluated for how accurately they capture real behavior~\citep{Kolluri2025-sv}.

Recent work has taken an important step in this direction by fine-tuning LLMs on Psych-101, a large-scale dataset of human decisions across diverse psychological experiments~\citep{Binz2024-vx,Binz2025-jo}. The resulting model, Centaur, predicts human behavior across a broad range of tasks and often outperforms task-specific cognitive models~\citep{Binz2025-jo}. Critically, however, Psych-101 does not link the same participants across multiple tasks. Centaur is therefore not directly trained to capture individual behavioral regularities across tasks or to use behavior in one task as evidence about the same individual in another.

This distinction matters because individual differences are central to cognitive science and realistic social simulation~\citep{Bryan2021-ya,Fisher2018-qk}. In cognitive science, cross-task prediction tests whether individual differences are task-specific or stable across tasks~\citep{Yarkoni2022-ty}. For example, can a participant's exploratory behavior in a bandit task or fairness preference in a social decision task inform behavior in other domains? Such regularities also matter for social simulation when outcomes depend on whether tendencies concentrate within the same individuals or are distributed across individuals, as in political sorting and clustered public-health risks~\citep{Baldassarri2008-da,Meader2016-kp}.

Here, we study cross-task prediction of individual behavior. Given a source task description, source-task behavior from a participant, and a target task description, the goal is to predict that participant's target-task behavior. To enable this evaluation, we collect a new dataset, Cross-Psych-101, in which the same participants complete multiple decision-making tasks. We then train Cross-Centaur on this cross-task prediction objective.

Results show that Cross-Centaur improves target-task action prediction from source-task behavior, including for held-out participants and for transfers between both related and distinct task categories.
Prompt ablations confirm that these gains depend on correctly aligned participant-specific source behavior rather than only on the target-task description or generic source-task context.
Probing analyses further suggest that Cross-Centaur's hidden states better capture interpretable behavioral tendencies in cross-task settings.
When visualized, participant representations from one task become organized by the target-task tendency, indicating the model transforms source-task evidence into target-relevant participant representations.

Our contributions are: (i) formulating cross-task individual behavior prediction as a supervised language-modeling problem, (ii) introducing Cross-Psych-101, a dataset in which the same participants complete multiple psychological decision-making tasks, and (iii) training and evaluating Cross-Centaur, showing that source-task behavior can improve target-task prediction and that the learned representations encode interpretable cross-task behavioral tendencies.

\section{Related Work}
\paragraph{LLMs for behavioral and social simulation.}
Large language models are increasingly used to simulate human behavior in experiments and agent-based simulations~\citep{Mou2026-rm,Gao2024-uc,Liu2025-vy}, often as synthetic participants or proxy agents~\citep{Aher2023-nr,Argyle2023-tw,Filippas2024-lx,Asano2024-ge}.
Related work populates interactive environments with believable agents~\citep{Park2023-ke,Gao2024-uc,Tang2025-fw,Mou2026-rm}.
These applications rely on the assumption that LLM agents are sufficiently faithful proxies for human behavior. Whether LLMs can provide such fidelity remains an empirical question.

\paragraph{Differences between LLMs and human decision-making.}
Recent studies show that LLMs can both reflect and differ from human biases, with issues like response instability, low psychometric validity, and moral judgment biases~\citep{Bisbee2024-nj,Dominguez-Olmedo2024-rt,Petrov2024-tx,Wang2025-sx,Kwon2025-ad,cheung2025amplified,Echterhoff2024-gm,Liu2025-tu,Coda-Forno2024-dm}.
One reason is that LLMs are trained to generate text, not to mimic human cognition or decision-making~\citep{Dillion2023-fh,Demszky2023-ez}.

\paragraph{Behavioral models and cross-task prediction.}
Cross-task prediction evaluates whether behavioral models generalize person-specific patterns: parameters from one task predict the same person's trial-level decisions in another~\citep{Ahn2008-zv,Robinson2023-gj}.
Neural methods instead learn low-dimensional individual representations from behavioral sequences~\citep{Dezfouli2019-lf}, including transfer across related task conditions~\citep{Higashi2025-jc}. LLM behavioral models have mainly learned from data pooled across experiments, as in Centaur and fine-tuning on social-science studies~\citep{Binz2025-jo,Kolluri2025-sv}. Other LLM individualization approaches use personas or self-reports, but can be incomplete and prompt-sensitive~\citep{Li2025-hx,Park2024-la,Hu2024-qg,Lutz2025-uu}.
We unify these approaches by training a language model to predict a participant's choices in a new task from their source-task trial-by-trial behavior.

\begin{figure*}[h]
    \centering
    \includegraphics[width=0.98\linewidth]{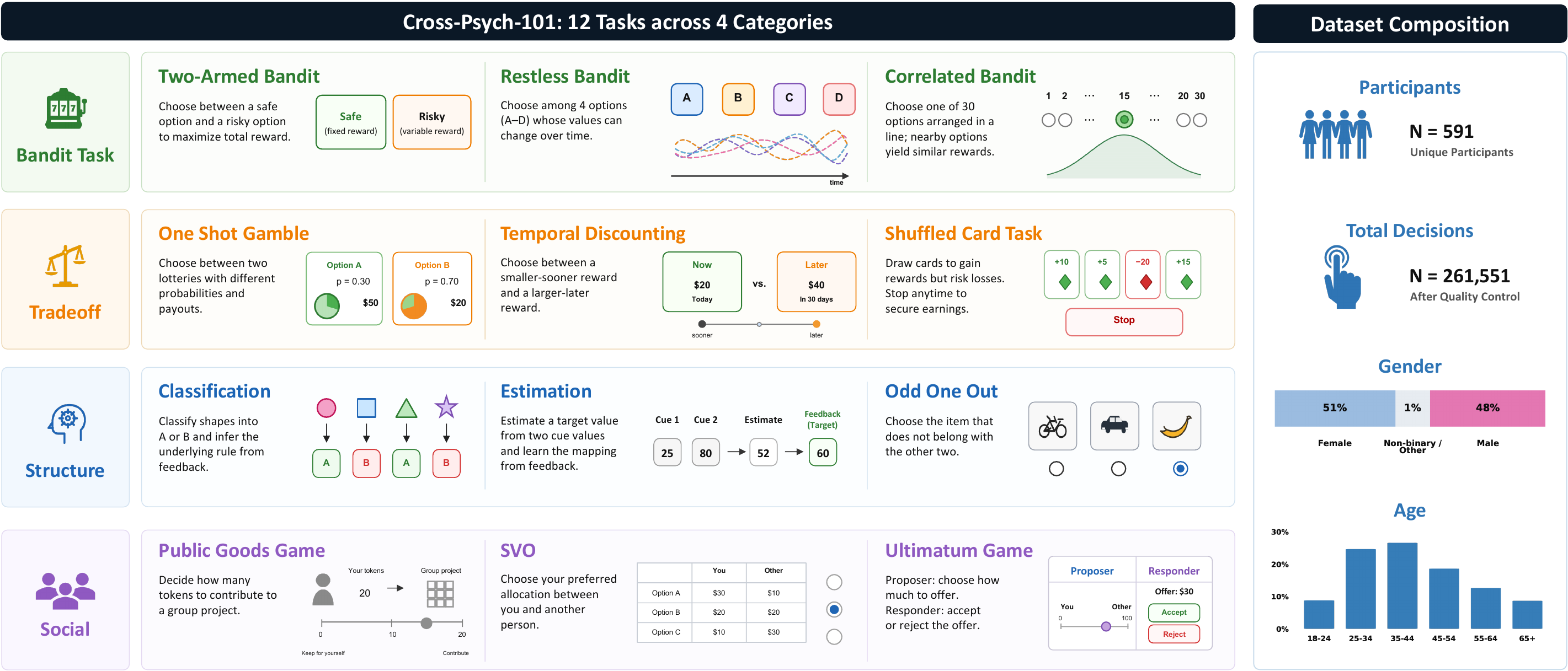}
    \caption{Overview of the Cross-Psych-101 dataset collection. The suite comprises twelve tasks spanning four domains (Bandit, Tradeoff, Structure, and Social). Each participant completes tasks from multiple domains, enabling analysis of individual behavioral signatures within and across domains.}
    \label{fig:dataset_overview}
\end{figure*}

\section{Task Formulation}
\label{sec:problem}

We formulate \emph{cross-task prediction of individual behavior}, which includes both same-task prediction and our main cross-task setting.
As shown in Fig.~\ref{fig:teaser}, each example consists of a \emph{source task} and a \emph{target task} completed by the same participant.

\subsection{Task and Behavior Representation}

Following prior work~\citep{Binz2024-vx,Binz2025-jo}, we represent each psychological task as a natural-language sequence.
Each task $k$ has a description $d_k$ specifying its rules, actions, and reward or feedback structure when applicable.
For participant $u$, the behavior transcript $\tau_{u,k}$ records their decisions and observed outcomes in task $k$.
A completed observation is therefore the pair $(d_k, \tau_{u,k})$.
A shortened task description example is shown below.
\begin{center}
    \begingroup
    \setlength{\fboxsep}{6pt}
    \setlength{\fboxrule}{0.4pt}
    \fcolorbox{black!25}{black!3}{
    \begin{minipage}{0.84\linewidth}
    \footnotesize\raggedright
    \textbf{Task description example}\\[2pt]
    \texttt{In this task, you are asked to choose between a safe option and a risky option.}\\
    \texttt{...}\\
    \texttt{Repeat your choices and select options to maximize your reward.}\\
    \end{minipage}
    }
    \endgroup
\end{center}

The behavior transcript uses the same textual format, with participant actions marked by \texttt{\textless\textless ...\textgreater\textgreater}.
These delimiters identify the action values used for supervised training and evaluation, as detailed in Section~\ref{sec:method}, reflecting our focus on action prediction rather than feedback generation or transcript surface form.
\begin{center}
    \begingroup
    \setlength{\fboxsep}{6pt}
    \setlength{\fboxrule}{0.4pt}
    \fcolorbox{black!25}{black!3}{
    \begin{minipage}{0.84\linewidth}
    \footnotesize\raggedright
    \textbf{Behavior transcript example}\\[2pt]
    \texttt{Chose \textless\textless risky\textgreater\textgreater{} and received -11 points.}\\
    \texttt{Chose \textless\textless safe\textgreater\textgreater{} and received 0 points.}\\
    \texttt{...}\\
    \texttt{Chose \textless\textless safe\textgreater\textgreater{} and got 0 points.}
    \end{minipage}
    }
    \endgroup
\end{center}

\subsection{Input--Output Formulation}

Each example is constructed from two task observations from the same participant.
Let $k_s$ and $k_t$ denote the source and target task indices.
The model input is
\[
x = (d_{k_s}, \tau_{u,k_s}, d_{k_t}).
\]

The prediction target is the same participant's target-task transcript:

\[
y = \tau_{u,k_t}.
\]

If $k_s = k_t$, this is same-task prediction: the model continues that task's transcript~\citep{Binz2024-vx,Binz2025-jo}.
If $k_s \neq k_t$, this is cross-task prediction, testing whether behavior in one task predicts another for the same participant.

\section{Cross-Psych-101 Construction}

We present Cross-Psych-101, a dataset for cross-task behavioral prediction across psychological domains. Unlike most behavioral datasets, in which participants complete a single task, Cross-Psych-101 follows the same participants across multiple decision-making tasks, yielding supervised examples as described in Section~\ref{sec:problem}.

\subsection{Task Selection and Design}

As shown in Fig.~\ref{fig:dataset_overview}, the task suite includes twelve tasks: three each from the \texttt{Bandit}, \texttt{Tradeoff}, \texttt{Structure}, and \texttt{Social} categories.
This setup enables evaluation of both \emph{within-category} (shared patterns) and \emph{across-category} (transfer of individual differences) generalization.

\paragraph{Task selection rationale.}
We chose these categories to cover key paradigms in human decision-making~\citep{Rangel2008-qr,Weber2009-dx}, while keeping the Psych-101 format~\citep{Binz2025-jo} for compatibility. Together, they represent complementary sources of individual variation:

\begin{itemize}
    \item \textbf{Bandit tasks}: measure how participants learn from reward feedback and balance exploration with exploitation~\citep{Gershman2018-cs, Whittle1988-fk, Schulz2020-fh}.
    \item \textbf{Tradeoff tasks}: measure how participants choose among options that differ in uncertainty, risk, reward magnitude, or delay~\citep{Frey2017-hw, Holt2002-op, van-den-Bos2013-zu}.
    \item \textbf{Structure tasks}: capture how participants infer categories, estimate continuous values, or make similarity- and category-based judgments over stimuli~\citep{Collins2014-fx, Flesch2018-sl, Hebart2023-mx}.
    \item \textbf{Social tasks}: examine how participants make decisions in socially interdependent settings, including cooperation, fairness, and preferences over allocations between themselves and others~\citep{Fischbacher2001-kc, Murphy2011-qi, Fehr1999-nc}.
\end{itemize}

\subsection{Collection Protocol}
\begin{figure}[t]
    \centering
    \includegraphics[width=1\linewidth]{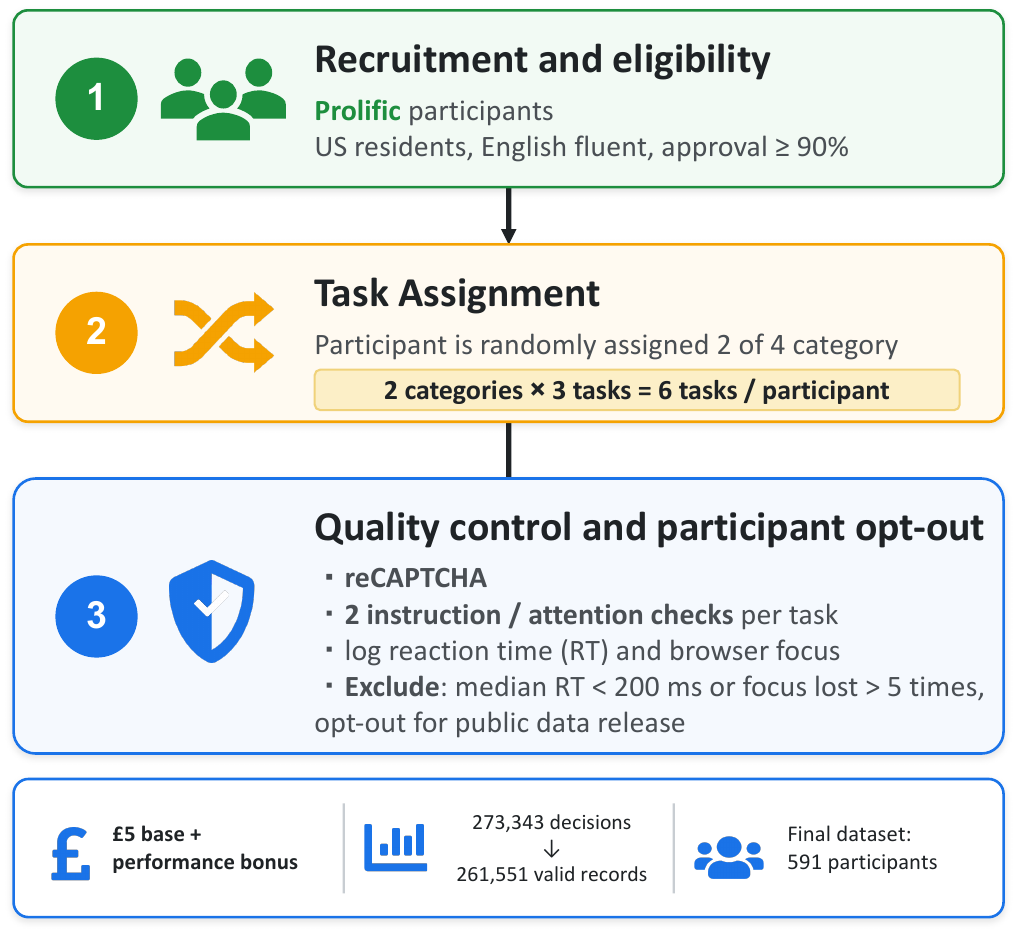}
    \caption{Summary of the data collection protocol. (1) Participants were recruited on Prolific with eligibility screening. (2) Each was randomly assigned two of four task categories and completed three tasks per category. (3) Quality control included reCAPTCHA, attention checks, response-time and browser-focus filters, and an opt-out for public release. The released dataset contains 261{,}551 valid decisions from 591 participants.}
    \label{fig:data-collection-protocol}
\end{figure}

Fig.~\ref{fig:data-collection-protocol} summarizes the data collection protocol, which consists of three steps: participant recruitment, task assignment, and quality control. We describe each step below.

\paragraph{Recruitment and eligibility.}
Participants were recruited through Prolific~\footnote{https://www.prolific.com/}, which offers a large and relatively high-quality participant pool and built-in screening mechanisms for AI-generated responses~\citep{Peer2022-fz}. Because the study was conducted in English, we restricted participation to US residents who self-reported being fluent in English and who had a prior Prolific approval rate of at least 90\%.

\paragraph{Task Assignment.}
Participants completed all tasks online on their own desktop or laptop computers. In each collection phase, each participant was randomly assigned two of the four task categories, with quotas used to keep the six category-pair combinations approximately balanced (See also Fig.~\ref{fig:task-cooccurrence-heatmap}). Participants completed the assigned categories in randomized order, while the task order within each category was fixed to reduce burden.

\paragraph{Collection design.}
This collection scheme made it possible to examine both generalization across different tasks within the same category and generalization across tasks in different categories, while keeping the burden per participant at a reasonable level. The resulting dataset provides enough coverage to study individual differences both within and across diverse experimental paradigms.

\begin{figure*}[t]
    \centering
    \includegraphics[width=0.98\linewidth]{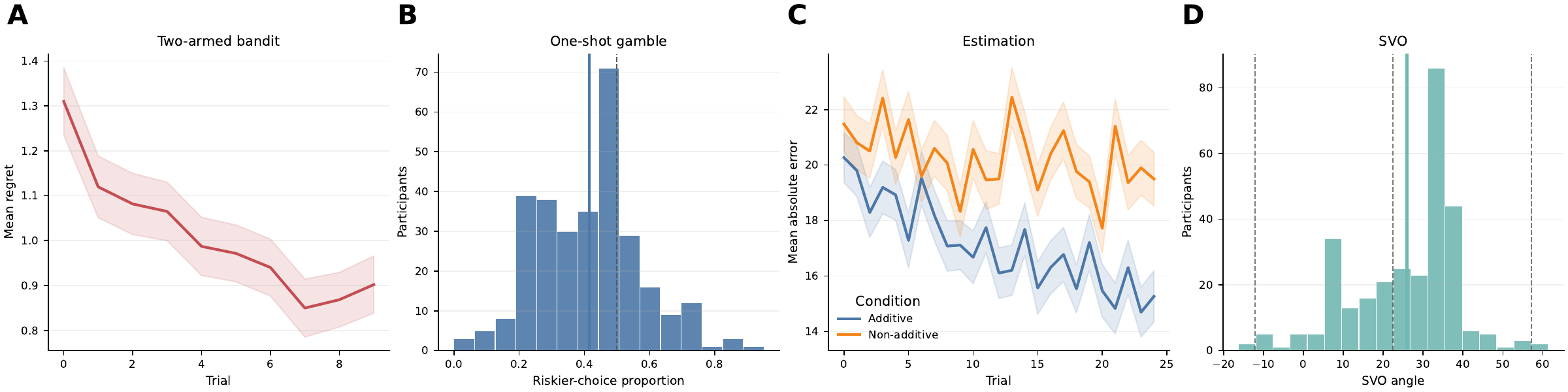}
    \caption{Validation analyses for four tasks, with one task selected from each category. Each panel summarizes participant-level performance or preferences in a different task. (A) Mean regret in the Two-Armed Bandit task. Shaded areas indicate standard errors. (B) Distribution of the proportion of trials in which participants chose the riskier lottery in the One-Shot Gamble task. (C) Mean absolute error in the Estimation task, shown separately for each target function. Shaded areas indicate standard errors. (D) Distribution of Social Value Orientation (SVO) scores. For the remaining eight tasks, see Appendix Fig.~\ref{fig:validation-8tasks}.}
    \label{fig:validation-4tasks}
\end{figure*}

\paragraph{Compensation.}\label{par:compensation}
Compensation consisted of a fixed base reward of \pounds 5.00 and a performance-based bonus. Performance-based bonuses were introduced to help maintain the motivation of participants throughout the task (see Appendix~\ref{sec:appendix-performance-based-bonus} for details). The median completion time was 36 minutes, corresponding to a mean effective hourly wage of approximately \pounds 8.19.

\paragraph{Ethics approval.}
The study protocol was approved by the ethics committee of the Graduate School of Humanities and Sociology, The University of Tokyo (UTSP-25051). All participants provided informed consent prior to participation, and no personally identifiable information will be released (see Appendix~\ref{sec:appendix-ethics} for further details).

\subsection{Quality Control and Participant Opt-Out}

To ensure data quality, we used several quality-control procedures. First, the experimental website used reCAPTCHA to prevent automated bot responses. Second, at the beginning of each task, participants read text-based instructions and completed two comprehension and attention-check questions before proceeding. Third, to sustain participants' motivation throughout the tasks, we introduced the performance-based bonus described in the previous section.

We logged browser interactions and millisecond-level response times for each task to detect inattention. Participant-task records were excluded if median response time was below 200 ms or if the browser window lost focus more than five times.
These exclusions were applied at the participant-task level, so participants could still provide valid data for other tasks.

Finally, independent of data quality, we respected participants' choices regarding data sharing. Before the public release, all participants could opt out of having their data included in the released dataset; data from those who opted out ($n = 10$) were excluded from the public release.

\section{Dataset Summary}

\begin{table*}[t]
    \centering
    \scriptsize
    \setlength{\tabcolsep}{3pt}
    \renewcommand{\arraystretch}{1.08}
    \resizebox{\textwidth}{!}{%
    \begin{tabular}{@{}l c cccc cccccccc@{}}
    \toprule
    \textbf{Model}
    & \textbf{Centaur}
    & \multicolumn{4}{c}{\textbf{Cross-Centaur Same-Task}}
    & \multicolumn{8}{c}{\textbf{Cross-Centaur Cross-Task}} \\
    \cmidrule(lr){2-2}
    \cmidrule(lr){3-6}
    \cmidrule(l){7-14}
    & Overall
    & Bandit & Tradeoff & Structure & Social
    & \multicolumn{2}{c}{Bandit}
    & \multicolumn{2}{c}{Tradeoff}
    & \multicolumn{2}{c}{Structure}
    & \multicolumn{2}{c}{Social} \\
    \cmidrule(lr){7-8}
    \cmidrule(lr){9-10}
    \cmidrule(lr){11-12}
    \cmidrule(l){13-14}
    &
    &
    &
    &
    &
    & Same cat. & Other cat.
    & Same cat. & Other cat.
    & Same cat. & Other cat.
    & Same cat. & Other cat. \\
    \midrule
    Qwen3-8B Base & 0.680 & 0.928 & 1.031 & 0.838 & 1.053 & 0.902 & 0.865 & 0.913 & 0.754 & 0.836 & 0.834 & 1.160 & 1.208 \\
    Qwen3-8B w/ reasoning & 1.009 & 1.681 & 1.522 & 1.421 & 1.552 & 1.531 & 1.159 & 1.267 & 0.959 & 1.265 & 1.294 & 1.620 & 1.563 \\
    Qwen3-8B w/ hints & 0.680 & 1.996 & 1.476 & 1.517 & 1.798 & 1.781 & 1.209 & 1.254 & 1.019 & 1.371 & 1.343 & 1.712 & 1.538 \\
    \midrule
    Qwen3-14B Base & 0.672 & 0.916 & 1.025 & 0.847 & 1.022 & 0.907 & 0.883 & 0.898 & 0.735 & 0.828 & 0.819 & 1.136 & 1.181 \\
    Qwen3-14B w/ reasoning & 0.953 & 1.735 & 1.399 & 1.206 & 1.586 & 1.579 & 1.278 & 1.224 & 0.934 & 1.178 & 1.226 & 1.504 & 1.530 \\
    Qwen3-14B w/ hints & 0.672 & 2.410 & 1.602 & 1.344 & 2.122 & 2.151 & 1.626 & 1.446 & 1.198 & 1.332 & 1.398 & 1.870 & 1.709 \\
    \midrule
    Centaur & 0.652 & 0.861 & 1.069 & 0.856 & 1.043 & 0.847 & 0.828 & 0.950 & 0.802 & 0.865 & 0.875 & 1.171 & 1.204 \\
    Cross-Centaur & \textbf{0.646} & \textbf{0.812} & \textbf{0.987} & \textbf{0.817} & \textbf{1.009} & \textbf{0.799} & \textbf{0.787} & \textbf{0.872} & \textbf{0.720} & \textbf{0.806} & \textbf{0.809} & \textbf{1.127} & \textbf{1.160} \\
    \bottomrule
    \end{tabular}%
    }
    \caption{Performance breakdown across Cross-Centaur evaluation settings, measured by negative log likelihood (lower is better). ``Same cat.'' and ``Other cat.'' denote whether the source task belongs to the same psychological category as the target task. For Centaur and Cross-Centaur, we report the average over three training seeds.}
    \label{tab:cross_centaur_breakdown}
    \end{table*}

\subsection{Final Dataset Statistics}

After applying the quality-control procedures described above, the full sample comprises 601 participants. Excluding the 10 participants who opted out of public data sharing, the publicly released dataset comprises 261{,}551 valid decisions from 591 participants (mean age = 42.2 years, standard deviation = 14.1; self-reported gender: 300 female, 285 male, and 6 other or not disclosed; Fig.~\ref{fig:dataset_overview}). Quality-based exclusions removed approximately 2.9\% of the 273{,}343 decisions originally recorded and were applied at the participant-task level rather than the participant level; thus, participants excluded from one task could still contribute valid data for other tasks.

\subsection{Decision Quality Analysis}

We inspected participant-level summaries for four representative tasks, one from each task category (Fig.~\ref{fig:validation-4tasks}), to confirm that the retained records reflected meaningful engagement. Specifically, we checked that participants responded to task structure in feedback-based tasks, and that preference tasks captured individual variation beyond ceiling or floor effects. Summary statistics for the remaining eight tasks are provided in Appendix~\ref{sec:appendix-decision-quality-analysis-details}.

\paragraph{Bandit.}
For feedback-based learning tasks, we assessed whether participants improved their choices over repeated trials. We defined regret as the difference between the expected score under the best available action and the realized score. In the Two-Armed Bandit task (Fig.~\ref{fig:validation-4tasks}A), mean regret decreased from $M = 1.31$ at trial~0 to $M = 0.90$ at trial~9. This decrease indicates that participants engaged with the task as intended and adjusted their choices in response to feedback.

\paragraph{Tradeoff.}
For tasks measuring preferences rather than objectively correct responses, we examined whether responses showed sufficient variability across participants. In the One-Shot Gamble task, each participant's proportion of riskier choices was distributed without strong skew (Fig.~\ref{fig:validation-4tasks}B), suggesting that the task captured meaningful individual variation rather than ceiling or floor effects.

\paragraph{Structure.}
For tasks that required inference from experience, we examined whether performance improved across trials and differed by task difficulty. In the Estimation task (Fig.~\ref{fig:validation-4tasks}C), participants' mean absolute error (MAE) decreased across trials. MAE was also consistently lower in the Additive condition ($\mathrm{MAE} = 15.3$) than in the Non-additive condition ($\mathrm{MAE} = 19.5$), consistent with the greater complexity of the non-additive target functions.

\paragraph{Social.}
For social-preference tasks, we assessed whether response distributions reproduced established behavioral patterns.
In the Social Value Orientation task, participants' scores were broadly concentrated in two regions corresponding to prosocial and individualistic preferences (Fig.~\ref{fig:validation-4tasks}D).
As this distribution has been consistently reported previously~\citep{Murphy2011-qi}, it further supports the validity of our data.

\paragraph{Summary.}
Together, these analyses indicate that Cross-Psych-101 reproduces canonical behavioral patterns across diverse psychological paradigms, supporting its suitability as a basis for the modeling analyses presented in the following sections.

\begin{table*}[t]
    \centering
    \footnotesize %
    \setlength{\tabcolsep}{3pt} %
    \renewcommand{\arraystretch}{1.09}
    \resizebox{1\linewidth}{!}{%
    \begin{tabular}{lcccccccccccc}
        \toprule
         & 2-Armed Bandit & Restless Bandit & Correlated Bandit & One-Shot Gamble & Temporal Discounting & Shuffled Card & Classification & Estimation & Odd One Out & Public Goods & Social Value & Ultimatum Game \\
        \midrule
        Qwen3-8B & 0.937 / 0.607 & 1.201 / 1.140 & 3.602 / 3.370 & 0.841 / 0.709 & 1.129 / 0.805 & 1.904 / 1.953 & 1.030 / 1.018 & 1.230 / 1.413 & 1.788 / 1.397 & 1.318 / 1.341 & 1.908 / 1.714 & 5.031 / 3.757 \\
        Qwen3-14B & 1.290 / 0.616 & 1.416 / 1.362 & 4.235 / 4.860 & 0.970 / 0.812 & 1.638 / 0.617 & 2.162 / 2.325 & 1.526 / 1.727 & 1.265 / 1.498 & 1.324 / 1.217 & 1.390 / 1.359 & 3.164 / 2.202 & 6.525 / 4.747 \\
        Centaur & 0.300 / 0.315 & 0.804 / 0.815 & 1.493 / 1.819 & \textbf{0.411} / \textbf{0.505} & 0.274 / 0.294 & \textbf{1.358} / 1.633 & 0.691 / 0.684 & 0.928 / 0.996 & \textbf{0.798} / \textbf{0.814} & \textbf{0.808} / 0.870 & \textbf{1.613} / \textbf{1.618} & 1.583 / 1.830 \\
        Cross-Centaur & \textbf{0.294} / \textbf{0.313} & \textbf{0.786} / \textbf{0.808} & \textbf{1.462} / \textbf{1.780} & \textbf{0.411} / 0.520 & \textbf{0.269} / \textbf{0.292} & 1.365 / \textbf{1.630} & \textbf{0.681} / \textbf{0.676} & \textbf{0.925} / \textbf{0.987} & 0.803 / 0.821 & 0.810 / \textbf{0.865} & 1.629 / 1.649 & \textbf{1.554} / \textbf{1.777} \\
        \bottomrule
    \end{tabular}%
    }
    \caption{Task-level negative log likelihood on Cross-Psych-101. Lower values are better. Each cell reports participant-held-out / task-held-out performance for the corresponding target task.}
    \label{tab:task_level_nll}
    \end{table*}

\begin{table}[h]
    \centering
    \footnotesize
    \setlength{\tabcolsep}{2.5pt}
    \begin{tabular}{lcccc}
        \toprule
            & Bandit & Tradeoff & Structure & Social \\
        \midrule
        Qwen3-8B / Shuffled & 1.731 & 1.463 & 1.393 & 2.115 \\
        Qwen3-8B / Target & 1.708 & 1.395 & \underline{1.318} & \underline{1.630} \\
        Qwen3-8B / Orig & \underline{1.689} & \underline{1.336} & 1.403 & 1.756 \\
        \midrule
        Centaur / Shuffled & 0.964 & 1.756 & 1.093 & 1.697 \\
        Centaur / Target & 0.969 & 1.899 & 1.125 & 1.635 \\
        Centaur / Orig & \underline{0.932} & \underline{1.580} & \underline{1.081} & \underline{1.612} \\
        \midrule
        Cross-Centaur / Shuffled & 0.798 & 0.858 & 0.786 & 1.119 \\
        Cross-Centaur / Target & 0.810 & 0.889 & 0.804 & 1.125 \\
        Cross-Centaur / Orig & \textbf{\underline{0.774}} & \textbf{\underline{0.790}} & \textbf{\underline{0.780}} & \textbf{\underline{1.065}} \\
        \bottomrule
    \end{tabular}
    \caption{Performance breakdown (negative log-likelihood, lower is better) by task family. Abbreviations: Orig.=Original prompt, Shuffled=Shuffled source, Target=Target only.}
    \label{tab:ablation_results}
    \end{table}

\section{Experiments}
\label{sec:experiments}

Our experiments address three main questions.
First, can a language model predict a participant's behavior in a target task from their behavior in a source task? (Sec.~\ref{sec:cross_task_prediction})
Second, does this generalize to unseen participants and to tasks withheld during training? (Sec.~\ref{sec:cross_task_prediction}; Sec.~\ref{sec:generalization})
Third, after behavioral fine-tuning, do the model's hidden states capture the cross-task tendencies of individual users? (Sec.~\ref{sec:probing})

\subsection{Training Formulation}
\label{sec:method}

Following Sec.~\ref{sec:problem}, for participant $u$ and source and target tasks $k_s$ and $k_t$, we represent each example as a chat-style supervised fine-tuning instance: the user message contains the source task description $d_{k_s}$, source behavior transcript $\tau_{u,k_s}$, and target task description $d_{k_t}$, and the assistant message is the target transcript $y=\tau_{u,k_t}$.
Thus, the model learns to generate a participant's target-task behavior from their source-task behavior. Appendix~\ref{app:prompt_template} shows the prompt template in detail.

Because we evaluate action prediction rather than transcript reconstruction, we compute loss only on the target action values.
Let $A(y)$ denote the token positions corresponding to the action values enclosed by \texttt{\textless\textless ...\textgreater\textgreater} in the target transcript, excluding the delimiter tokens themselves.
For a model $p_\theta$, we minimize
\[
\mathcal{L}(\theta)
=
-
\sum_{(x,y)\in\mathcal{D}}
\sum_{i \in A(y)}
\log p_\theta(y_i \mid x, y_{<i}),
\]
where $y_i$ is the $i$-th token in the target transcript.
Tokens outside the marked action values, such as feedback, outcome text, and delimiters, are excluded from the loss.
This ensures that training targets participants' decision-making, rather than peripheral aspects of the environment or generic instructional text.

\subsection{Experimental Setup}

We use Qwen3-8B~\citep{Yang2025-yi} as the base model.
We first fine-tune it on Psych-101~\citep{Binz2025-jo} in our chat-style action-prediction template to obtain \textbf{Centaur}, then adapt it to Cross-Psych-101 as \textbf{Cross-Centaur}.
In Cross-Psych-101, the model predicts target-task behavior from a source task description, the participant's source-task behavior, and a target task description.

We compare \textbf{Qwen3-8B} and \textbf{Qwen3-14B} directly and with two prompt modifications: explicit reasoning before prediction and quantitative summaries of prior behavioral tendencies.
These conditions test whether such cues help base models capture cross-task patterns. For more details, see Appendix~\ref{app:training_details}.
To test robustness to the choice of base model, we additionally train and evaluate Centaur and Cross-Centaur with different model families and sizes (Qwen3-1.7B~\citep{Yang2025-yi}, Llama-3.2-1B, and Llama-3.1-8B~\citep{Grattafiori2024-lm}); see Appendix~\ref{app:model_variants} for these results.

\subsection{Metrics}
Following prior work~\citep{Binz2024-vx,Binz2025-jo}, we use \emph{negative log likelihood (NLL)} as our primary evaluation metric.
NLL evaluates how well a model predicts behavioral actions, with lower values indicating better performance.
It is calculated only over action values enclosed by \texttt{\textless\textless ...\textgreater\textgreater}, excluding delimiters and other tokens.

\subsection{Dataset Splits}
\label{sec:dataset_splits}

We use two splits for complementary evaluation purposes.\footnote{All reported experiments use the full collected sample of 601 participants, prior to the public-release opt-out. The publicly released dataset comprises 591 participants.}
The \emph{participant-held-out} split assigns approximately 80\% of participants to training and 20\% to testing, and is used for the cross-task prediction and prompt-ablation analyses in Sec.~\ref{sec:cross_task_prediction}.
The \emph{task-held-out} split excludes certain tasks from training entirely (as both source and target), so train/test sizes depend on the data for each held-out task; this split is used for the generalization analysis in Sec.~\ref{sec:generalization}.
These splits together test generalization to unseen participants and tasks.

\subsection{Cross-Task Behavioral Prediction}
\label{sec:cross_task_prediction}

Table~\ref{tab:cross_centaur_breakdown} summarizes action-token negative log likelihood across same-task and cross-task settings.
In this table, the Centaur and Cross-Centaur rows report the mean over three training seeds.
All other rows, e.g., Qwen3-8B Base, do not involve any training, and NLL is deterministic given the model and test data; therefore, they are unaffected by random seeds.\footnote{Standard deviations across the training seeds in Table~\ref{tab:cross_centaur_breakdown} are reported in Appendix~\ref{app:seed_variability}. In addition, although NLL is deterministic for a fixed model, we report bootstrap comparisons over the test data among Qwen3-8B Base, Centaur, and Cross-Centaur in Appendix~\ref{app:bootstrap_analysis}.}
The Cross-Centaur model achieves the best score in nearly all columns, improving over both off-the-shelf Qwen baselines and the Psych-101-tuned Centaur model.
Explicit reasoning prompts and quantitative behavioral hints also fail to improve the off-the-shelf Qwen models, indicating that these gains are not obtained simply from inference-time cues.
This shows that Cross-Psych-101 fine-tuning provides information beyond a general language prior and beyond population-level behavioral alignment from Psych-101.

The cross-task columns further separate source tasks from the same psychological category as the target task from those in other categories.
Same-category transfer tests whether individual differences generalize across related paradigms, while other-category transfer asks whether behavior in one domain helps predict another.
Cross-Centaur's improvements in both settings support direct tests of reusable individual behavioral signatures across tasks, which are not possible in single-task behavioral datasets.

\begin{table*}[t]
    \centering
    \scriptsize
    \setlength{\tabcolsep}{3pt}
    \renewcommand{\arraystretch}{1.08}
    \resizebox{\textwidth}{!}{%
    \begin{tabular}{@{}l cccc cccccccc@{}}
    \toprule
    \textbf{Model}
    & \multicolumn{4}{c}{\textbf{Same-Task}}
    & \multicolumn{8}{c}{\textbf{Cross-Task}} \\
    \cmidrule(lr){2-5}
    \cmidrule(l){6-13}
    & Bandit & Tradeoff & Structure & Social
    & \multicolumn{2}{c}{Bandit}
    & \multicolumn{2}{c}{Tradeoff}
    & \multicolumn{2}{c}{Structure}
    & \multicolumn{2}{c}{Social} \\
    \cmidrule(lr){6-7}
    \cmidrule(lr){8-9}
    \cmidrule(lr){10-11}
    \cmidrule(l){12-13}
    &
    &
    &
    &
    & Same cat. & Other cat.
    & Same cat. & Other cat.
    & Same cat. & Other cat.
    & Same cat. & Other cat. \\
    \midrule
    Qwen3-8B & 0.724 & 0.759 & 0.567 & \textbf{0.603} & 0.583 & 0.567 & 0.531 & 0.438 & 0.533 & 0.533 & 0.596 & \textbf{0.574} \\
    Centaur & \textbf{0.738} & 0.759 & 0.600 & \textbf{0.603} & 0.600 & 0.556 & 0.542 & 0.438 & 0.533 & 0.533 & 0.596 & 0.569 \\
    Cross-Centaur & 0.734 & \textbf{0.760} & \textbf{0.633} & \textbf{0.603} & \textbf{0.634} & \textbf{0.609} & \textbf{0.749} & \textbf{0.477} & \textbf{0.567} & \textbf{0.632} & \textbf{0.608} & 0.553 \\
    \bottomrule
    \end{tabular}%
    }
    \caption{Probe accuracy breakdown by psychological category on Cross-Psych-101. Higher values are better. ``Same cat.'' and ``Other cat.'' denote whether the source task belongs to the same psychological category as the target task.}
    \label{tab:category_nll_probe_sweep}
    \end{table*}

\subsubsection{Prompt Ablations}
\label{sec:prompt_ablations}

We further test whether Cross-Centaur's improvements are due to participant-specific source-task evidence, rather than the target task description or generic source-task context.
Table~\ref{tab:ablation_results} compares three prompt conditions on the participant-held-out split.
Here, $d_k$ denotes the description of task $k$, and $\tau_{u,k}$ denotes participant $u$'s behavior transcript in that task.
For the same target $y=\tau_{u,k_t}$, we vary the input defined in Sec.~\ref{sec:problem} as follows:
\[
\begin{aligned}
x_{\mathrm{Orig.}} &= (d_{k_s}, \tau_{u,k_s}, d_{k_t}),\\
x_{\mathrm{Shuffled}} &= (d_{k_s}, \tau_{u',k_s}, d_{k_t}), \quad u' \neq u,\\
x_{\mathrm{Target}} &= (d_{k_t}).
\end{aligned}
\]
Thus, \emph{Orig.} provides matched source behavior from the same participant, \emph{Shuffled} preserves the source-task format but breaks participant alignment, and \emph{Target} removes the source observation entirely.

Across task families, the original prompt is strongest for Cross-Centaur.
The degradation under \emph{Shuffled} indicates that mismatched participant information is not sufficient and can be harmful, while the degradation under \emph{Target} shows that source-task behavior provides predictive information beyond the target task description alone.
Together, these ablations support the interpretation that Cross-Centaur uses participant-specific cross-task evidence rather than only task-level priors or prompt length effects.

\subsection{Task-Level Generalization}
\label{sec:generalization}

Table~\ref{tab:task_level_nll} compares Centaur and Cross-Centaur at the task level, reporting participant-held-out and task-held-out scores for each target task.
The task-held-out setting tests whether cross-task adaptation transfers to target paradigms absent from Cross-Psych-101 fine-tuning.
Cross-Centaur improves over Centaur and Qwen3-8B in the task-held-out setting, indicating that it can generalize to unseen target paradigms.
However, these gains are smaller than those in Table~\ref{tab:cross_centaur_breakdown}, where the target task family is represented during fine-tuning.
A similar gap between seen and unseen target tasks is also observed for Centaur, suggesting that further improving generalization to entirely unseen task types remains an important direction for future work.

\begin{figure}[t]
    \centering
    \includegraphics[width=\columnwidth]{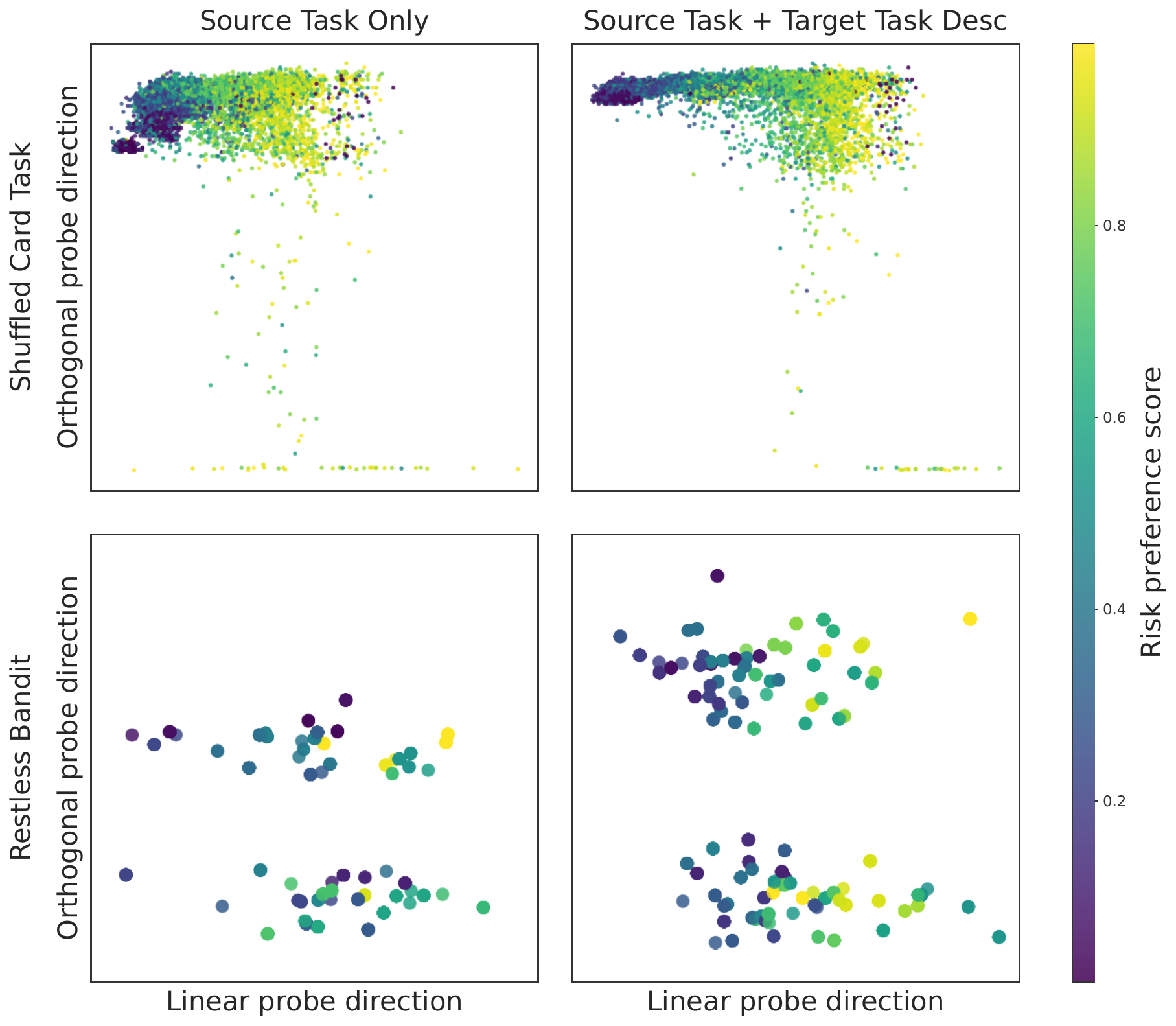}
    \caption{
        Cross-Centaur hidden states visualized by shuffled-card-task risk-preference score.
        Each point represents a participant, colored by their observed risk preference (yellow = more risk-seeking, blue = more risk-averse).
        The horizontal axis shows the hidden state’s projection onto a linear probe for this score, so more risk-seeking participants appear further right; the vertical axis is orthogonal and just spreads points in two dimensions.
        Rows vary the source task (top: shuffled card, same task; bottom: restless bandit, cross task), and columns compare inputs with only source-task behavior (left) versus those also including the shuffled-card target-task description (right).
        Adding the target description clarifies the left-to-right risk preference gradient, especially for the cross-task (bottom) row.
    }
    \label{fig:internal_representation}
\end{figure}

\subsection{Probing Behavioral Representations}
\label{sec:probing}

\paragraph{Probing Behavioral Tendencies}

To test whether the models encode participant-level behavioral tendencies, we use diagnostic probing to assess whether Cross-Centaur’s hidden states have organized behavioral evidence into representations from which individual differences can be decoded~\citep{Alain2016-yx,Belinkov2022-yv,Liu2019-me}.
For each participant-task transcript, we convert choices and outcomes into task-specific tendency scores that summarize behavior, \eg regret in Bandit tasks or risk preference in Tradeoff tasks (see Appendix~\ref{app:tendency_scores}).
We extract hidden states from each model and train single-hidden-layer MLP probes to predict these scores, using a separate probe for each target task. See Appendix~\ref{app:probing_setup} for more details.

Table~\ref{tab:category_nll_probe_sweep} shows that same-task probing performance is often comparable across models, suggesting that even off-the-shelf language models can partially represent behavioral tendencies when the input comes from the same task.
In contrast, Cross-Centaur yields stronger probe accuracy in cross-task settings, indicating that its hidden states better preserve participant-specific information that transfers across tasks.
This supports the idea that Cross-Centaur captures richer participant representations, not just task-specific patterns.

\paragraph{Internal Representations}

Figure~\ref{fig:internal_representation} complements the probe-accuracy results with a direct visualization of Cross-Centaur's hidden states, using the shuffled-card-task risk-preference score (Appendix~\ref{app:tendency_scores}) as the target tendency.
For each prompt, we extract the mean-pooled hidden state as in the probing setup (Appendix~\ref{app:probing_setup}) and fit a linear probe that predicts the participant's risk-preference score from it.
Each point is a participant, placed by the projection of their hidden state onto the fitted probe direction (horizontal axis) and onto an orthogonal direction (vertical axis), and colored by their observed risk preference.
Thus, the horizontal position reflects how risk-seeking the model's internal representation encodes the participant to be, and a smooth left-to-right color gradient indicates that the tendency is linearly decodable from the hidden states; the vertical axis serves only to spread the points.

The top row uses the shuffled card task itself as the source.
These same-task representations are already organized by risk preference from source behavior alone (top left), and the gradient becomes cleaner once the target-task description is appended (top right), suggesting that specifying the target task helps the model express participant information in a form aligned with the relevant behavioral tendency.
The bottom row is the cross-task case, where the source is the restless bandit task.
From restless-bandit behavior alone (bottom left), the representations are less cleanly organized by shuffled-card risk preference, as expected because the input has not yet specified the target domain.
Once the shuffled-card target description is appended (bottom right), participants become more clearly organized by their observed shuffled-card tendency, even though the model has seen no shuffled-card behavior from them.
Together with Cross-Centaur's stronger cross-task probe accuracy in Table~\ref{tab:category_nll_probe_sweep}, this pattern is consistent with the model using source-task behavior to form participant representations that are aligned with the specified target task.

\section{Conclusion}

We studied cross-task prediction of individual behavior, in which a participant's behavior in one psychological task is used to predict their behavior in another.
To this end, we introduced Cross-Psych-101, a dataset in which the same participants complete multiple decision-making tasks, and trained Cross-Centaur on this cross-task prediction objective.
Cross-Centaur improves target-task prediction over off-the-shelf language models and Centaur, including for held-out participants and for transfers across task categories, and prompt ablations confirm that these gains depend on participant-specific source behavior.
Probing and visualization analyses further suggest that its hidden states encode behavioral tendencies that transfer across tasks.
We hope that Cross-Psych-101 and Cross-Centaur serve as a foundation for future work toward unified models of individual human decision-making.

\section*{Limitations}

Although Cross-Psych-101 and Cross-Centaur provide a step toward cross-task modeling of individual behavior, the scope of the present study remains limited. A long-term goal in psychological modeling is to build a unified behavioral model that can reproduce and evaluate human decision-making across a broad range of situations~\citep{Binz2024-vx,Binz2025-jo}. In this sense, Cross-Psych-101 is still restricted to controlled psychological tasks. The dataset covers multiple decision-making domains, including bandit, tradeoff, structure, and social tasks, but it does not yet capture the full range of decisions people make in everyday, institutional, or socially embedded settings. The participant pool is also linguistically limited: because the study was conducted in English, data were collected from English-fluent participants, which may limit generalization to speakers of other languages and to cultural contexts where the same task descriptions may be interpreted differently. Future datasets should connect psychological decision-making tasks with more general social and real-world decision contexts and should include broader linguistic and cultural coverage.

There are also limitations in the training formulation. Our results show that constructing a cross-task dataset enables models to use source-task behavior for target-task prediction, and that Cross-Centaur can acquire task-transferrable representations of participants. However, the gains are more modest in the task-held-out setting, and the probing results are limited in some cases. The additional regret analysis in Appendix~\ref{app:two_armed_bandit_human_model_regret} further illustrates this limitation: Cross-Centaur captures the overall participant-level tendency in Two-armed bandit regret, but its predicted regret distribution is narrower than the human distribution. Thus, while the current model provides evidence for cross-task individual representations, it does not fully solve the problem of learning robust, general-purpose representations that preserve the full diversity of human decision-making.

Future work may extend this line of work by using training objectives that more explicitly encourage representation learning across tasks. Another promising direction is to extract participant characteristics from source-task behavior as intermediate natural-language reasoning, and then use those inferred characteristics to guide target-task prediction. More broadly, our work should be viewed as an initial step toward the larger goal of a unified model for psychology, rather than a complete solution.

\section*{Ethical Considerations and Potential Risks}

Models grounded in human behavioral data should not be interpreted as complete models of human society. Even when such models reproduce observed behavioral patterns, human decision-making is shaped by complex social, cultural, institutional, and historical contexts that are difficult to capture in a finite dataset or model. Therefore, LLM-based behavioral simulations should be treated as tools for exploring possible patterns or hypotheses, not as sources of definitive answers about individuals or society. Any use of these models should recognize that their outputs are conditional on the data, tasks, and modeling assumptions used to construct them, and should not replace human judgment or domain expertise in consequential settings.

\section*{Acknowledgments}
This work was supported by the International Graduate Program of Innovation for Intelligent World at the University of Tokyo. This research was conducted as part of the Value Exchange Engineering project, a collaborative effort between R4D, Mercari, Inc., and RIISE.
Additional financial support was provided by the Japan Society for the Promotion of Science (JP26KJ0793 to H.S. and JP23KJ0781 to R.M.), JST SPRING (JPMJSP2108 to H.S.), and JST ACT-X (JPMJAX25CQ to R.M.).

\bibliography{custom}

\appendix

\section{Contribution Details}
\label{sec:contribution}

This appendix details the specific contributions of each author.

\paragraph{Hikaru Asano} proposed the main idea of Cross-Centaur and led the modeling and experimental work. He formulated the cross-task individual-level prediction approach, implemented and trained Cross-Centaur, and designed and carried out the experiments, including the comparisons against off-the-shelf LLMs and Centaur on held-out participants and the probing analyses examining whether the model's hidden states encode participant-level behavioral tendencies that transfer across tasks.

\paragraph{Hidezo Suganuma} led the design of the Cross-Psych-101 dataset and built the data collection pipeline. This involved selecting and adapting twelve tasks across four categories, designing the cross-task collection protocol (random category assignment, balanced quotas, and performance-based bonuses), and implementing the experimental website for data collection on Prolific, including quality-control mechanisms such as reCAPTCHA, comprehension and attention checks, and response-time logging.

\paragraph{Ryutaro Mori} contributed to the main idea of Cross-Centaur and to the design of the Cross-Psych-101 dataset. He helped shape the cross-task individual-level prediction formulation underlying Cross-Centaur and the dataset design, including the choice of task categories and the cross-task collection protocol. He was also involved in quality control of the collected data and in the analyses of participant-level behavioral tendencies used to validate the dataset across task categories.

\paragraph{Yukino Baba} As supervisor, she provided general guidance and advice throughout the research process. She advised on the experimental design and the interpretation of results as needed, and supervised the overall direction of the project.

\section{Data Collection Details}

This section supplements the main text with further details on data collection and analyses of data quality.

\subsection{Ethical Considerations}\label{sec:appendix-ethics}

The data collection protocol was approved by the ethics committee of the Graduate School of Humanities and Sociology, The University of Tokyo (UTSP-25051). All participants provided informed consent prior to participation and were assigned anonymized participant identifiers; no personally identifiable information was retained beyond the pseudonymous IDs issued by Prolific. Prior to the public release, participants were additionally given the opportunity to opt out of data sharing, and data from those who opted out were excluded from the released dataset. These procedures were implemented to ensure ethical data handling while enabling public release of the dataset for research transparency and reproducibility.

\subsection{Experimental Interfaces}

\begin{figure*}[t]
    \centering
    \includegraphics[width=\textwidth]{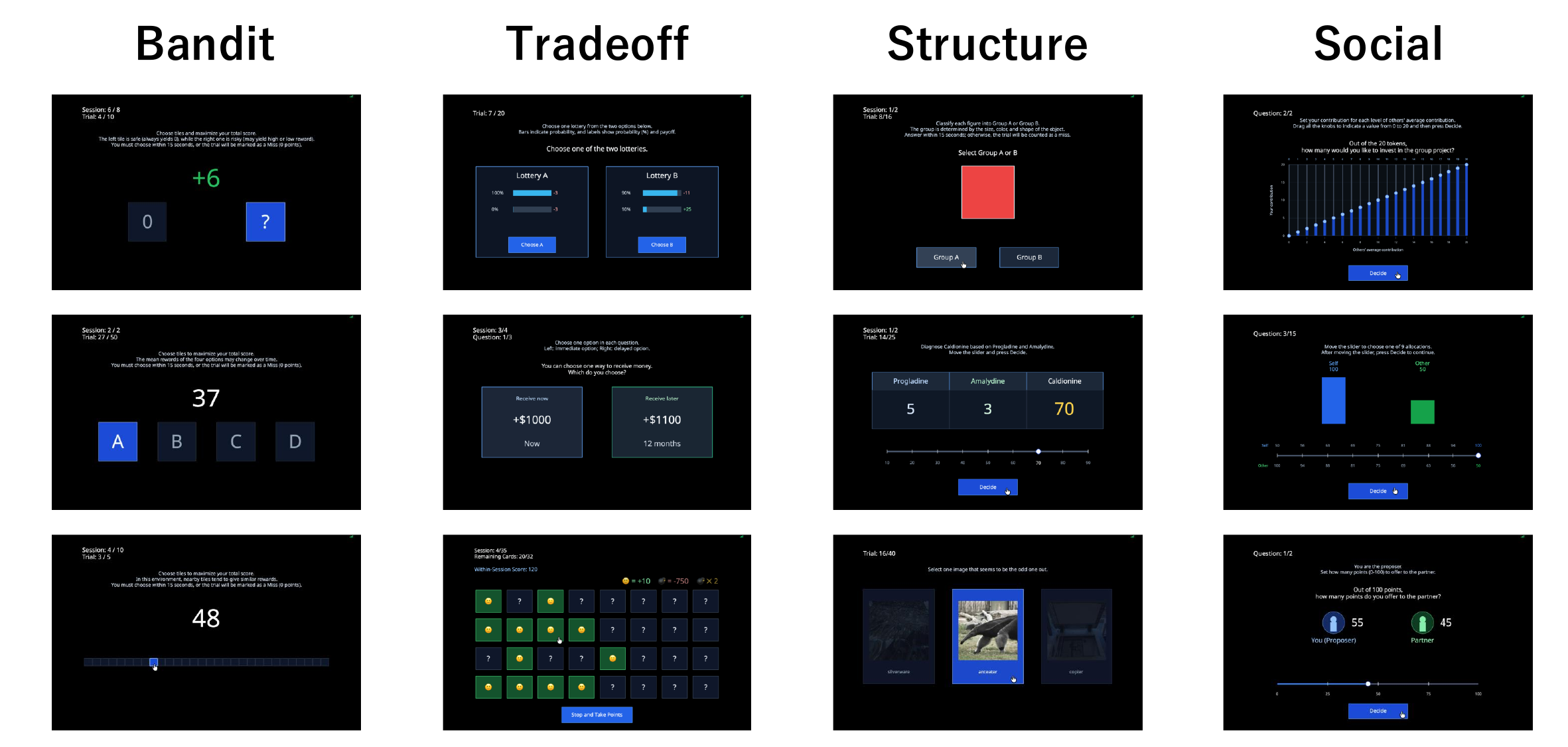}
    \caption{
    Screenshots of the experimental interfaces for the 12 tasks. Each task is classified under one of the following four categories: Bandit, Tradeoff, Structure, or Social.
    }
    \label{fig:screenshots_agg}
\end{figure*}

Participants completed the experiments through simple web interfaces, as shown in Fig.~\ref{fig:screenshots_agg}.

\begin{figure*}[t]
    \centering
    \includegraphics[width=0.98\linewidth]{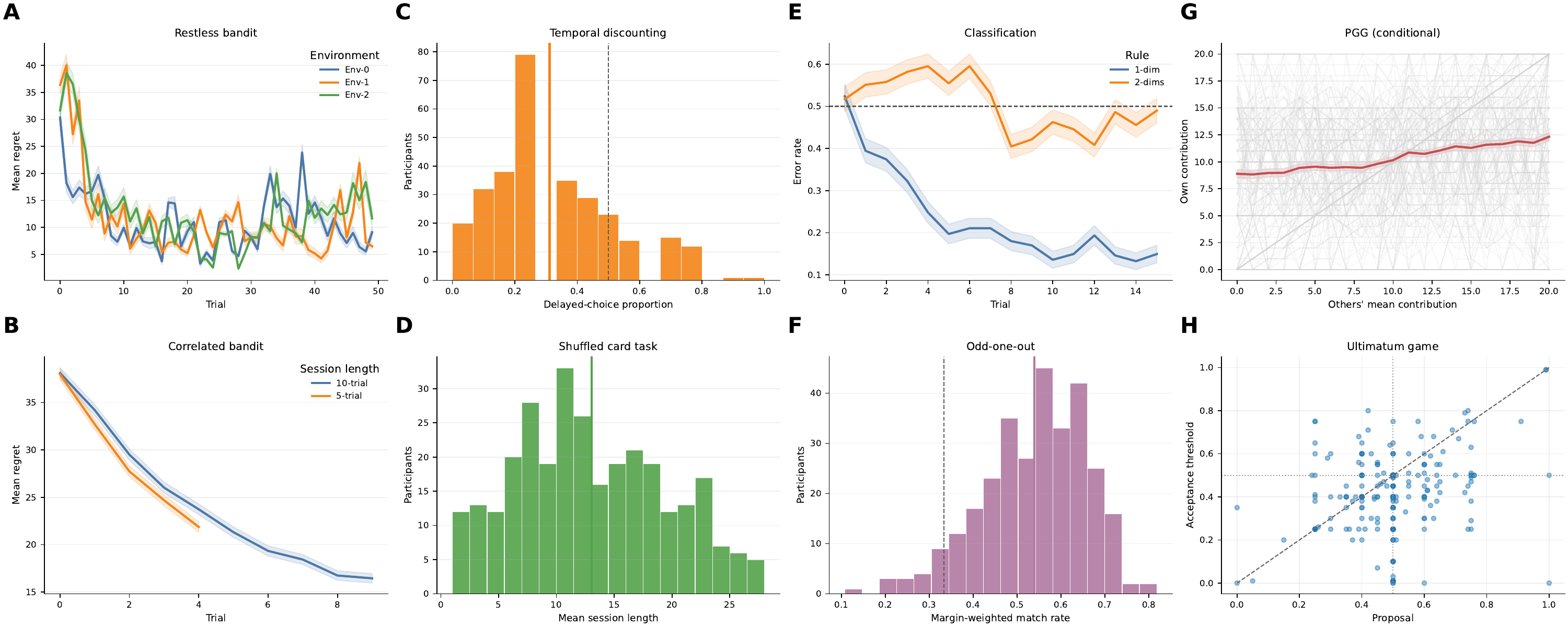}
    \caption{Supplementary validation analyses across eight tasks not discussed in the main text. Panels (A) and (B) show mean regret in the Restless and Correlated bandit tasks, respectively. Panel (C) shows the distribution of the proportion of trials in which participants chose the delayed option in the Temporal Discounting task. Panel (D) shows the distribution of mean session length, a proxy for risk preference, in the Shuffled Card task. Panel (E) shows mean error rate in the Classification task, separately for each classification rule. Shaded areas indicate standard errors. Panel (F) shows the margin-weighted match rate in the Odd-One-Out task, which can be interpreted as the consistency between participants’ subjective similarity judgments and distances derived from pretrained word-embedding vectors. Panel (G) shows mean contributions in the conditional Public Goods Game task. Panel (H) shows a scatterplot of each participant’s proposal and minimum acceptance threshold in the Ultimatum Game task.}
    \label{fig:validation-8tasks}
\end{figure*}

\subsection{Task Co-occurrence Balance}
\label{sec:appendix-task-cooccurrence-balance}

\begin{figure}[h]
    \centering
    \includegraphics[width=1\columnwidth]{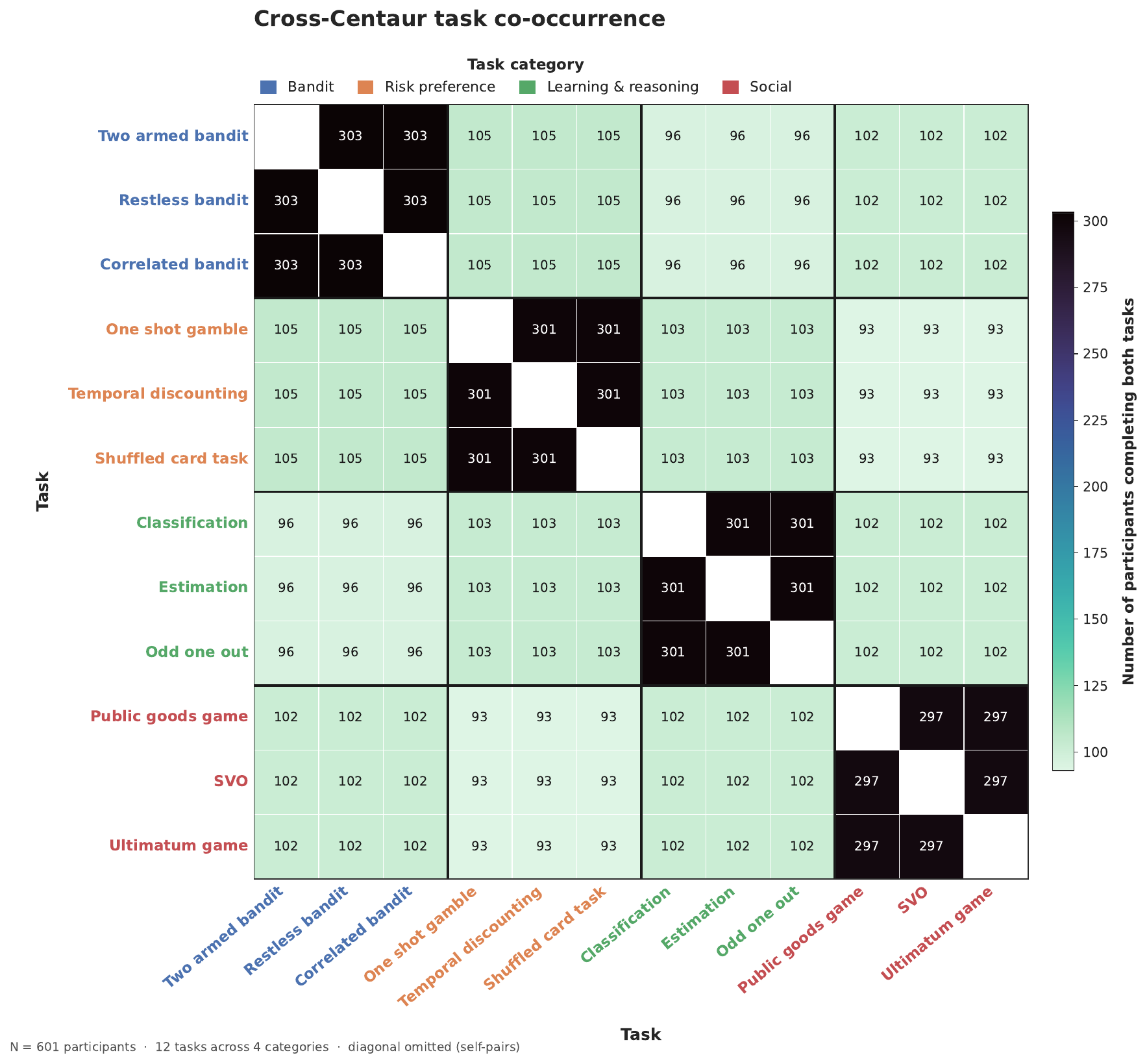}
    \caption{Task co-occurrence counts across participants. Each cell shows the number of participants who completed both tasks in the corresponding unordered task pair.}
    \label{fig:task-cooccurrence-heatmap}
\end{figure}

To verify that the cross-task combinations were approximately balanced, we visualized the task co-occurrence structure across participants. For each participant, we reconstructed the set of completed tasks from the processed data of the full experimental sample used in Section~\ref{sec:experiments} (601 participants; see Section~\ref{sec:dataset_splits}) and incremented the count for every unordered pair of tasks completed by that participant. The resulting symmetric co-occurrence matrix is shown in Fig.~\ref{fig:task-cooccurrence-heatmap}.

The block-diagonal structure reflects the experimental design: tasks within the same category were always administered together. Accordingly, within-category task pairs were observed for nearly all participants in that category: 303 participants for Bandit tasks, 301 for Tradeoff tasks, 301 for Structure tasks, and 297 for Social tasks. More importantly, the off-diagonal blocks show that cross-category task combinations were also well covered. Each task pair between two categories had identical counts within that category-pair block, indicating that participants were assigned at the category level rather than producing uneven task-specific pairings. The cross-category counts ranged from 93 to 105 participants per task pair: Bandit-Tradeoff = 105, Bandit-Structure = 96, Bandit-Social = 102, Tradeoff-Structure = 103, Tradeoff-Social = 93, and Structure-Social = 102.

Thus, although not every participant completed every task, the cross-task design provides broad and approximately balanced coverage across task categories. This balance supports cross-task analyses by reducing the possibility that observed cross-task effects are driven by a small number of overrepresented task-pair combinations.

\subsection{Performance-Based Bonuses}\label{sec:appendix-performance-based-bonus}

Performance-based bonuses in the present experiment were administered according to the following procedure, taking into account their potential influence on task performance. As described in Sec.~\ref{par:compensation}, participants received a fixed base reward of \pounds 5.00 and could receive an additional performance-based bonus. After data collection, one of the 9 incentivized tasks listed in Table~\ref{tab:task-list} was selected at random. For the selected task, the top 50 participants, ranked by accumulated points, or, for error-based tasks, by the lowest error magnitude, received an additional bonus of \pounds 3.00.
Thus, participants who received the bonus earned \pounds 8.00 in total.

Participants collectively received approximately 677{,}000 yen in rewards, with an additional one-third of this amount paid as fees to Prolific; thus, the total expenditure for dataset collection was roughly 903{,}000 yen.

\subsection{Additional Analysis on Data Quality}\label{sec:appendix-decision-quality-analysis-details}

This section reports visualizations of summary statistics of participants' responses for the eight tasks not discussed in the main text. Figs.~\ref{fig:validation-8tasks}A and ~\ref{fig:validation-8tasks}B show participants' mean regret in the Restless bandit and Correlated bandit tasks, respectively, presented in the same format as the Two-armed bandit task; both panels indicate that participants learned from experience. Fig.~\ref{fig:validation-8tasks}C shows the proportion of delayed-option choices in the Temporal Discounting task, and Fig.~\ref{fig:validation-8tasks}D shows the distribution of session continuation counts in the Shuffled Card task. Both distributions are approximately normal, suggesting that the tasks adequately captured the diversity of participants' temporal discounting and risk preferences. Fig.~\ref{fig:validation-8tasks}E presents the error rates in the Classification task, which, as in the Estimation task, exhibited the predicted pattern of responses across conditions of varying difficulty. For the Odd-One-Out task, for instance, we computed the margin-weighted match rate between participants' choices and the option identified as most ``odd'' based on the embedding vectors of the word stimuli, obtained using the all-MiniLM-L6-v2 model~\citep{Reimers2019-iy}. The match rate was well above the chance level of $1/3$ ($M = 0.54$), indicating that participants' responses were systematically related to the semantic structure of the stimuli rather than being random. The remaining two Social-category tasks also captured the diversity of participants' responses. As shown in Figs.~\ref{fig:validation-8tasks}G and ~\ref{fig:validation-8tasks}H, the overall trend in conditional contributions and the prevalence of equal-split proposals in the Ultimatum game are both patterns that have been robustly reported in the prior literature~\citep{Fischbacher2001-kc, Henrich2001-fu, Guth1982-px}. 

Overall, these results indicate that the collected dataset reproduces canonical behavioral patterns reported in prior psychological experiments across diverse paradigms, supporting the suitability of the data as a basis for the modeling analyses.

\begin{table*}[h]
    \centering
    \scriptsize 
    \setlength{\tabcolsep}{3pt}
    \renewcommand{\arraystretch}{1.08}
    \resizebox{\textwidth}{!}{%
    \begin{tabular}{@{} p{3cm}p{2.2cm}p{2.2cm}p{3cm}p{8cm} @{}}
        \toprule
        Task Name & Category & Incentivized? & Reference & Summary \\
        \midrule
        Two-armed bandit & Bandit & Yes & \citet{Gershman2018-cs} & Choose between a safe option (always yielding zero points) and a risky option (yielding a positive or negative reward). 10 trials $\times$ 8 sessions; the risky option's reward distribution varies across sessions. \\
        \addlinespace
        Restless bandit & Bandit & Yes & \citet{Daw2006-ws} & Choose one of four options whose expected rewards drift over time. 50 trials $\times$ 2 sessions. \\
        \addlinespace
        Correlated bandit & Bandit & Yes & \citet{Wu2018-fz} & Choose one of thirty options arranged in a row, where neighboring options yield similar rewards. 10 sessions in total (5 long sessions of 10 trials and 5 short sessions of 5 trials); reward distributions and scales vary across sessions. \\
        \addlinespace
        One-shot gamble & Tradeoff & Yes & \citet{Krueger2024-od} & Choose between two lotteries, each specified by probabilities and payoffs, for 20 trials. One chosen lottery is resolved at the end of the task. \\
        \addlinespace
        Temporal discounting & Tradeoff & No & \citet{Ruggeri2022-pf} & Choose between an immediate (sooner) and a delayed (later) monetary option across varying amounts and delays. 3 trials $\times$ 4 sessions, covering both gain and loss conditions. \\
        \addlinespace
        Shuffled card task & Tradeoff & Yes & \citet{Frey2017-hw} & Repeatedly draw a card from a deck of 32 (containing gain and bomb cards) or stop. Drawing a bomb card forfeits the session's accumulated points and ends the session. 35 sessions, varying in gain amount, loss amount, and number of bomb cards. \\
        \addlinespace
        Classification & Structure & Yes & \citet{Speekenbrink2008-fh} & Classify figures (varying in shape, color, and size) into category A or B based on feedback. 16 trials $\times$ 2 sessions, each with a different classification rule. \\
        \addlinespace
        Estimation & Structure & Yes & \citet{Collsioo2023-nu} & Estimate a target value (a fictional hormone, 10-90) from two numerical cues (1-5). 25 trials $\times$ 2 sessions; the mapping from cues to the target is either additive (easy) or non-additive (hard). \\
        \addlinespace
        Odd one out & Structure & No & \citet{Hebart2023-mx} & Choose the odd one out among three images. 40 trials in a single session; designed to measure conceptual similarity rather than to have a correct answer. \\
        \addlinespace
        Public goods game & Social & Yes & \citet{Fischbacher2001-kc} & Decide how many of 20 tokens to contribute to a group project shared by four members (including oneself). Two sessions: unconditional contribution (1 trial, without information about others' contributions) and conditional contribution (21 trials, one for each possible average of the others). \\
        \addlinespace
        Social value orientation & Social & No & \citet{Murphy2011-qi} & Choose the preferred allocation of money between self and an anonymous other from 9 options. 15 trials in a single session. \\
        \addlinespace
        Ultimatum game & Social & Yes & \citet{Henrich2001-fu} & Split 100 points with another participant. Each participant acts as both proposer (deciding how many points to offer) and responder (deciding the minimum acceptable offer). \\
        \bottomrule
    \end{tabular}
    }
    \caption{List of the 12 tasks used in the experiment. This table details the category, the presence of financial incentives, the reference, and the specific task description for each task. Note that due to time constraints, the number of trials and sessions has generally been reduced compared to the original tasks.}
    \label{tab:task-list}
\end{table*}

\section{Additional Training Details}
\label{app:training_details}

This section provides implementation details for Cross-Centaur fine-tuning and the representation probing analyses.
We first describe the fine-tuning setup and prompt format, and then report the probing configuration and behavioral tendency scores used as probe targets.

\subsection{Fine-Tuning Setup}
\label{app:fine_tuning_setup}

All Cross-Centaur models were initialized from Qwen3-8B.
We first fine-tuned Qwen3-8B\footnote{\url{https://huggingface.co/Qwen/Qwen3-8B}}, which is distributed under the Apache License 2.0, on Psych-101\footnote{\url{https://huggingface.co/datasets/marcelbinz/Psych-101}} to obtain the Centaur checkpoint. Both resources were utilized in accordance with their respective licenses and intended use policies. Subsequently, this checkpoint was adapted to Cross-Psych-101 using the chat-style supervised fine-tuning format shown in Fig.~\ref{fig:prompt_fullwidth}.

Model training was implemented with Unsloth~\citep{unsloth} to improve the efficiency of supervised fine-tuning.
All training, evaluation, and probing experiments were run on a machine with eight NVIDIA GH200 Grace Hopper Superchip GPUs.
The fine-tuning hyperparameters are summarized in Table~\ref{tab:fine_tuning_hyperparameters}.

For the participant-held-out split, 80\% of participants were assigned to the training side and 20\% to the test side.
For task-held-out experiments, the same hyperparameters were reused after removing the held-out task from both the source and target sides of the Cross-Psych-101 training examples.

\begin{table*}[t]
    \centering
    \small
    \setlength{\tabcolsep}{5pt}
    \renewcommand{\arraystretch}{1.08}
    \begin{tabular}{@{}p{3.8cm}p{11.5cm}@{}}
        \toprule
        Setting & Value \\
        \midrule
        Base model & Qwen3-8B \\
        Training framework & Unsloth\\
        Hardware & Eight NVIDIA GH200 Grace Hopper Superchip GPUs \\
        Trainable parameters & LoRA adapter weights only \\
        LoRA configuration & Rank $r=16$, $\alpha=32$, dropout $0.05$, no bias terms \\
        LoRA target modules & \texttt{q\_proj}, \texttt{k\_proj}, \texttt{v\_proj}, \texttt{o\_proj}, \texttt{gate\_proj}, \texttt{up\_proj}, \texttt{down\_proj} \\
        Objective & Action-only language-modeling loss on tokens inside marked action spans \\
        Optimizer and schedule & AdamW with a linear learning-rate schedule \\
        Learning rate & $2\times10^{-6}$ \\
        Number of epochs & 1 \\
        Batch and accumulation & Per-device batch size 8, gradient accumulation step 1 \\
        Maximum gradient norm & 1.0 \\
        \bottomrule
    \end{tabular}
    \caption{Fine-tuning hyperparameters for the Qwen3-8B-based Cross-Centaur models.}
    \label{tab:fine_tuning_hyperparameters}
\end{table*}

\subsection{Probing Setup}
\label{app:probing_setup}

For the representation probing analysis, we used the Qwen3-8B, Centaur, and Cross-Centaur variants compared in the base experiments.
The goal of this analysis was to test whether model hidden states encode participant-level behavioral tendencies that can be decoded by a lightweight diagnostic model.
For each Cross-Psych-101 example, we rendered only the user-side prompt, which contains the source task description, the participant's source-task behavior transcript, and the target task description.

We then extracted hidden states from this prompt without including the assistant target transcript, so that the probe used only information available before target-task behavior was generated.
As described in Sec.~\ref{sec:probing}, we trained separate single-hidden-layer MLP probes for the task-specific tendency scores listed in Table~\ref{tab:tendency_scores}.
For the probe-accuracy results, each numeric tendency score was binarized by the median value in the corresponding target column, and the probe predicted whether the participant's score was above that median.
This setup was applied to both same-task inputs, where the source and target tasks were identical, and cross-task inputs, where the source task differed from the target task.
The probing hyperparameters are summarized in Table~\ref{tab:probing_hyperparameters}.

For the visualization in Fig.~\ref{fig:internal_representation}, we additionally fit a linear probe on the same mean-pooled hidden states to predict the shuffled-card-task risk-preference score.
Each representation was then projected onto the fitted probe weight direction, shown as the horizontal axis, and onto a direction orthogonal to it, shown as the vertical axis.
Only the horizontal coordinate is informative about the decoded tendency; the orthogonal coordinate is included solely to display the points in two dimensions.

\begin{table*}[t]
    \centering
    \small
    \setlength{\tabcolsep}{5pt}
    \renewcommand{\arraystretch}{1.08}
    \begin{tabular}{@{}p{3.8cm}p{11.5cm}@{}}
        \toprule
        Setting & Value \\
        \midrule
        Model variants & Qwen3-8B, Centaur, and Cross-Centaur \\
        Representation input & User-side prompt only; assistant target transcript excluded \\
        Hidden-state layer & Final transformer layer \\
        Pooling & Mean pooling over non-padding tokens \\
        Model inference precision & 4-bit model loading with bfloat16 computation \\
        Probe scope & Separate probe for each task \\
        Probe model & One-hidden-layer MLP with 128 hidden units \\
        Activation & ReLU \\
        Optimizer & AdamW \\
        Learning rate & $10^{-3}$ \\
        Weight decay & $10^{-4}$ \\
        Probe batch size & 64 \\
        Number of epochs & 200 \\
        Evaluation metric & Binary classification accuracy \\
        \bottomrule
    \end{tabular}
    \caption{Hyperparameters for extracting model representations and training behavioral tendency probes.}
    \label{tab:probing_hyperparameters}
\end{table*}

\subsection{Behavioral Tendency Scores for Probing}
\label{app:tendency_scores}

For the probing analysis in Section~\ref{sec:experiments}, we computed participant-level tendency scores.
Each row corresponds to one participant (\texttt{p\_id}), and each numeric score column summarizes a representative behavioral tendency or performance measure for one task.
These numeric columns were used as prediction targets for MLP probes and as color variables when visualizing the model's internal representations.
Table~\ref{tab:tendency_scores} lists the score names used in the probing analysis.

\begin{table*}[t]
    \centering
    \scriptsize
    \setlength{\tabcolsep}{3pt}
    \renewcommand{\arraystretch}{1.08}
    \resizebox{\textwidth}{!}{%
    \begin{tabular}{@{}p{2.8cm}p{4.2cm}p{2.2cm}p{7.0cm}p{4.4cm}@{}}
        \toprule
        Task & Score Name & Type & Definition & Interpretation of Larger Values \\
        \midrule
        Two-armed bandit & \texttt{regret} & performance & Negative mean regret, where regret is best reward minus realized reward. & Better choices. \\
        \midrule
        Restless bandit & \texttt{regret} & performance & Negative mean regret, where regret is the maximum mean reward on each trial minus the realized reward. & Better choices. \\
        \midrule
        Correlated bandit & \texttt{regret} & performance & Negative mean regret, where regret is \texttt{rewardScale} $\times \max(\mu)$ minus the realized reward. & Better choices. \\
        \midrule
        One-shot gamble & \texttt{risk preference} & risk preference & Fraction of the 20 trials in which the participant chose the higher-variance option. & More risk-seeking choices. \\
        \midrule
        Shuffled card tas & \texttt{risk preference} & risk preference & Number of sessions ending by drawing a bomb, divided by 35. & More sessions in which the participant continues drawing until a bomb. \\
        \midrule
        Shuffled card task & \texttt{risk preference} & risk preference & Mean stopping trial among voluntarily stopped sessions, divided by 35. & Later stopping and stronger tendency to keep drawing. \\
        \midrule
        Classification & \texttt{classification accuracy} & performance & Mean classification accuracy, computed as the mean of \texttt{outcome}. & Higher accuracy. \\
        \midrule
        Estimation & \texttt{estimation error} & performance & Negative mean absolute estimation error. & Lower estimation error. \\
        \midrule
        Public goods game & \texttt{unconditional contribution} & social behavior & Unconditional contribution in session 0, ranging from 0 to 20. & Greater unconditional contribution. \\
        \midrule
        Public goods game & \texttt{behavioral type} & categorical & Conditional-contribution pattern in session 1: \texttt{FR} for free rider, \texttt{UC} for unconditional cooperation, \texttt{CC} for conditional cooperation, \texttt{TC} for triangular contribution, and \texttt{Other} for other patterns. & Not applicable because this is not a numeric score. \\
        \midrule
        Social value orientation & \texttt{SVO angle} & social behavior & Murphy-style SVO angle in degrees. & More prosocial or altruistic value orientation. \\
        \midrule
        Ultimatum game & \texttt{proposed share} & social behavior & Fraction of the endowment offered when acting as proposer, ranging from 0 to 1. & Larger proposed share for the other player. \\
        \midrule
        Ultimatum game & \texttt{acceptance threshold} & social behavior & Minimum acceptable offer when acting as responder, expressed as a fraction from 0 to 1. & Higher acceptance threshold. \\
        \bottomrule
    \end{tabular}%
    }
    \caption{Participant-level behavioral tendency scores used as targets for probing and as color variables for representation visualization.}
    \label{tab:tendency_scores}
\end{table*}

\begin{figure*}[t]
    \centering
    \includegraphics[width=0.92\textwidth]{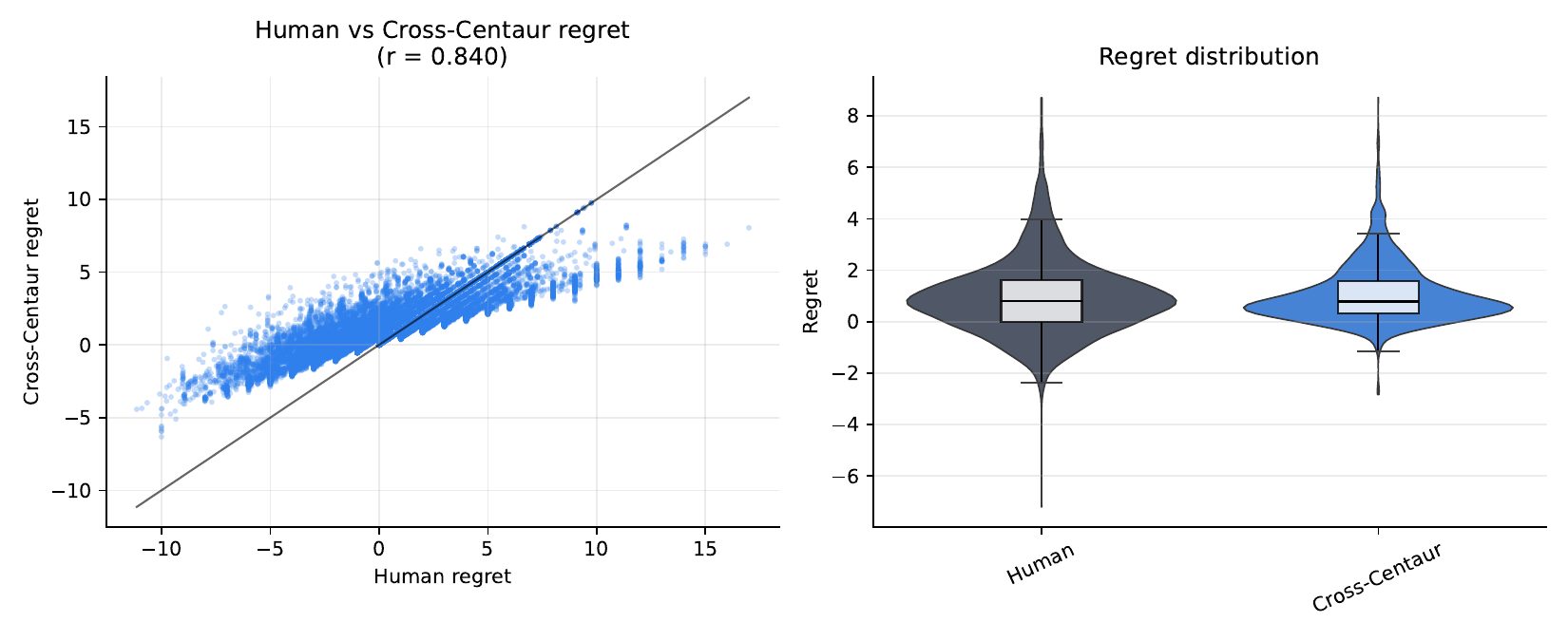}
    \caption{
    Comparison between human and Cross-Centaur regret in the Two-armed bandit task.
    The scatterplot shows participant-level regret values for humans and Cross-Centaur predictions, with a correlation of $r = 0.840$.
    The distribution plot shows that human regret is more broadly distributed than Cross-Centaur regret.
    }
    \label{fig:two_armed_bandit_summary}
\end{figure*}

\section{Additional Analysis}

\subsection{Human-Model Regret Patterns}
\label{app:two_armed_bandit_human_model_regret}

We further examined whether Cross-Centaur reproduces participant-level variation in the Two-armed bandit task.
Fig.~\ref{fig:two_armed_bandit_summary} compares each participant's human regret with the regret obtained from Cross-Centaur predictions.
The two quantities are strongly associated ($r = 0.840$), indicating that Cross-Centaur captures a substantial component of the individual differences expressed in human bandit behavior.
This result supports the central contribution of Cross-Centaur: by using behavior observed in one domain to condition predictions in another, the model can recover meaningful cross-domain tendencies in participants' decision-making.

At the same time, the distributional comparison in Fig.~\ref{fig:two_armed_bandit_summary} shows that human regret is more widely dispersed than Cross-Centaur regret.
Thus, although Cross-Centaur learns a robust signal about which participants tend to make higher- or lower-regret decisions, its predictions do not fully reproduce the breadth of human behavioral diversity.
One possible explanation is that additional model training can reduce output diversity, as discussed by \citet{Song2024-qj,Li2025-dz}, which may cause fine-tuned models to concentrate predictions around more typical behavioral patterns.
This remaining gap is an important limitation of the current model and points to a direction for future work: learning representations that preserve not only the central tendency of cross-task behavior but also the full heterogeneity of human decision-making.

\subsection{Prompt Ablations Details}

Table~\ref{tab:cross_centaur_breakdown_appendix} provides the full version of the ablation results summarized in Table~\ref{tab:ablation_results}.
Whereas the main text reports the compact category-level comparison, this appendix table retains the complete breakdown across same-task and cross-task settings, including whether the source task belongs to the same psychological category as the target task.

\begin{table*}[t]
    \centering
    \scriptsize
    \setlength{\tabcolsep}{3pt}
    \renewcommand{\arraystretch}{1.08}
    \resizebox{\textwidth}{!}{%
    \begin{tabular}{@{}l cccc cccccccc@{}}
    \toprule
    \textbf{Model}
    & \multicolumn{4}{c}{\textbf{Cross-Centaur Same-Task}}
    & \multicolumn{8}{c}{\textbf{Cross-Centaur Cross-Task}} \\
    \cmidrule(lr){2-5}
    \cmidrule(l){6-13}
    Bandit & Tradeoff & Structure & Social
    & \multicolumn{2}{c}{Bandit}
    & \multicolumn{2}{c}{Tradeoff}
    & \multicolumn{2}{c}{Structure}
    & \multicolumn{2}{c}{Social} \\
    \cmidrule(lr){6-7}
    \cmidrule(lr){8-9}
    \cmidrule(lr){10-11}
    \cmidrule(l){12-13}
    &
    &
    &
    &
    Same cat. & Other cat.
    & Same cat. & Other cat.
    & Same cat. & Other cat.
    & Same cat. & Other cat. \\
    \midrule
    Qwen3-8B-prompt Shuffled source & 2.076 & 1.822 & 1.431 & 2.569 & 1.837 & 1.281 & 1.488 & 1.080 & 1.368 & 1.379 & 2.031 & 1.746 \\
    Qwen3-8B-prompt Target only & 1.907 & 1.709 & 1.374 & 1.638 & 1.751 & 1.465 & 1.432 & 1.044 & 1.303 & 1.277 & 1.628 & 1.624 \\
    Qwen3-8B Original & 2.007 & 1.587 & 1.458 & 1.744 & 1.798 & 1.262 & 1.350 & 1.071 & 1.371 & 1.379 & 1.783 & 1.741 \\
    \midrule
    centaur prompt Shuffled source & 1.071 & 2.218 & 1.146 & 1.709 & 0.987 & 0.834 & 1.761 & 1.291 & 1.078 & 1.055 & 1.691 & 1.691 \\
    centaur prompt Target only & 1.097 & 2.482 & 1.223 & 1.584 & 0.991 & 0.819 & 1.918 & 1.296 & 1.097 & 1.055 & 1.648 & 1.673 \\
    centaur Original & 1.013 & 1.869 & 1.119 & 1.516 & 0.951 & 0.831 & 1.579 & 1.290 & 1.071 & 1.054 & 1.632 & 1.687 \\
    \midrule
    cross-centaur-both prompt Shuffled source & 0.834 & 1.040 & 0.809 & 1.103 & 0.804 & 0.756 & 0.877 & 0.658 & 0.777 & 0.771 & 1.130 & 1.125 \\
    cross-centaur-both prompt Target only & 0.863 & 1.096 & 0.829 & 1.108 & 0.822 & 0.745 & 0.913 & 0.658 & 0.797 & 0.785 & 1.129 & 1.138 \\
    cross-centaur-both Original & 0.792 & 0.910 & 0.795 & 0.982 & 0.776 & 0.754 & 0.802 & 0.656 & 0.773 & 0.771 & 1.094 & 1.120 \\
    \bottomrule
    \end{tabular}%
    }
    \caption{Performance breakdown across Cross-Centaur evaluation settings. Lower values are better. ``Same cat.'' and ``Other cat.'' denote whether the source task belongs to the same psychological category as the target task.}
    \label{tab:cross_centaur_breakdown_appendix}
\end{table*}

\subsection{Variability Across Training Seeds}
\label{app:seed_variability}

Table~\ref{tab:cross_centaur_breakdown} in Sec.~\ref{sec:cross_task_prediction} presents the Centaur and Cross-Centaur results averaged over three training seeds, while Table~\ref{tab:seed_variability} reports the corresponding means and standard deviations across seeds.
The standard deviations are small in all settings (at most 0.029, and below 0.005 in most columns) and are clearly smaller than the differences between Centaur and Cross-Centaur, indicating that the improvements reported in the main text are stable across training seeds.

\begin{table*}[t]
    \centering
    \scriptsize
    \setlength{\tabcolsep}{3pt}
    \renewcommand{\arraystretch}{1.08}
    \resizebox{\textwidth}{!}{%
    \begin{tabular}{@{}l c cccc cccccccc@{}}
    \toprule
    \textbf{Model}
    & \textbf{Centaur}
    & \multicolumn{4}{c}{\textbf{Cross-Centaur Same-Task}}
    & \multicolumn{8}{c}{\textbf{Cross-Centaur Cross-Task}} \\
    \cmidrule(lr){2-2}
    \cmidrule(lr){3-6}
    \cmidrule(l){7-14}
    & Overall
    & Bandit & Tradeoff & Structure & Social
    & \multicolumn{2}{c}{Bandit}
    & \multicolumn{2}{c}{Tradeoff}
    & \multicolumn{2}{c}{Structure}
    & \multicolumn{2}{c}{Social} \\
    \cmidrule(lr){7-8}
    \cmidrule(lr){9-10}
    \cmidrule(lr){11-12}
    \cmidrule(l){13-14}
    &
    &
    &
    &
    &
    & Same cat. & Other cat.
    & Same cat. & Other cat.
    & Same cat. & Other cat.
    & Same cat. & Other cat. \\
    \midrule
    Centaur & $0.652 \pm 0.002$ & $0.861 \pm 0.003$ & $1.069 \pm 0.019$ & $0.856 \pm 0.001$ & $1.043 \pm 0.002$ & $0.847 \pm 0.003$ & $0.828 \pm 0.002$ & $0.950 \pm 0.011$ & $0.802 \pm 0.000$ & $0.865 \pm 0.000$ & $0.875 \pm 0.001$ & $1.171 \pm 0.001$ & $1.204 \pm 0.001$ \\
    Cross-Centaur & $0.646 \pm 0.002$ & $0.812 \pm 0.002$ & $0.987 \pm 0.029$ & $0.817 \pm 0.001$ & $1.009 \pm 0.000$ & $0.799 \pm 0.002$ & $0.787 \pm 0.003$ & $0.872 \pm 0.017$ & $0.720 \pm 0.001$ & $0.806 \pm 0.001$ & $0.809 \pm 0.001$ & $1.127 \pm 0.002$ & $1.160 \pm 0.001$ \\
    \bottomrule
    \end{tabular}%
    }
    \caption{Variability of Centaur and Cross-Centaur across training seeds. Each cell reports the mean $\pm$ standard deviation of negative log likelihood over the three training seeds used in Table~\ref{tab:cross_centaur_breakdown} (lower is better). ``Same cat.'' and ``Other cat.'' denote whether the source task belongs to the same psychological category as the target task.}
    \label{tab:seed_variability}
    \end{table*}

\subsection{Bootstrap Significance Analysis}
\label{app:bootstrap_analysis}

Negative log likelihood (NLL) is deterministic for a fixed model and test set because it is computed directly from the predicted probabilities, without any sampling at inference time.
Nevertheless, to quantify the uncertainty of model comparisons, we performed a nonparametric bootstrap over the test data, resampling the participants in the test set 10{,}000 times with replacement and evaluating all models on the identical resamples.
We compare Qwen3-8B Base, Centaur, and Cross-Centaur, and report
$\Delta\mathrm{NLL} = \mathrm{NLL}_{\text{baseline}} - \mathrm{NLL}_{\text{Cross-Centaur}}$
with 95\% confidence intervals; positive values favor Cross-Centaur.
As shown in Table~\ref{tab:bootstrap_analysis}, all confidence intervals exclude zero, indicating that Cross-Centaur's improvements over both baselines are statistically reliable on Psych-101 as well as in the same-task and cross-task settings of Cross-Psych-101.

\begin{table}[t]
    \centering
    \footnotesize
    \setlength{\tabcolsep}{3pt}
    \resizebox{\columnwidth}{!}{%
    \begin{tabular}{lccc}
        \toprule
        & & \multicolumn{2}{c}{\textbf{Cross-Psych-101}} \\
        \cmidrule(l){3-4}
        \textbf{Baseline} & \textbf{Psych-101} & Same-Task & Cross-Task \\
        \midrule
        Qwen3-8B Base & 1.350 [1.232, 1.473] & 1.254 [1.051, 1.478] & 0.899 [0.666, 1.199] \\
        Centaur & 0.008 [0.003, 0.014] & 0.110 [0.070, 0.170] & 0.150 [0.084, 0.241] \\
        \bottomrule
    \end{tabular}%
    }
    \caption{Bootstrap comparison against Cross-Centaur. Each cell reports $\Delta\mathrm{NLL} = \mathrm{NLL}_{\text{baseline}} - \mathrm{NLL}_{\text{Cross-Centaur}}$ with 95\% confidence intervals from 10{,}000 participant-level bootstrap resamples of the test data; positive values favor Cross-Centaur.}
    \label{tab:bootstrap_analysis}
\end{table}

\subsection{Robustness Across Model Families and Sizes}
\label{app:model_variants}

To test whether our findings depend on the choice of base model, we fine-tuned additional backbone models spanning both model scale and model family: Qwen3-1.7B and Qwen3-8B~\citep{Yang2025-yi}, and Llama-3.2-1B and Llama-3.1-8B~\citep{Grattafiori2024-lm}.
For each backbone, we trained both Centaur and Cross-Centaur and evaluated them on the participant-held-out split.
As shown in Table~\ref{tab:model_variants}, Cross-Centaur outperforms Centaur in both the same-task and cross-task settings across all backbones.
Within each family, larger models also yield better performance, indicating that the cross-task objective benefits from scale and is not specific to a single model family.

\begin{table*}[t]
    \centering
    \scriptsize
    \setlength{\tabcolsep}{3pt}
    \renewcommand{\arraystretch}{1.08}
    \resizebox{\textwidth}{!}{%
    \begin{tabular}{@{}l cccc cccc@{}}
    \toprule
    \textbf{Model}
    & \multicolumn{4}{c}{\textbf{Same-Task}}
    & \multicolumn{4}{c}{\textbf{Cross-Task}} \\
    \cmidrule(lr){2-5}
    \cmidrule(l){6-9}
    & Bandit & Tradeoff & Structure & Social
    & Bandit & Tradeoff & Structure & Social \\
    \midrule
    Llama-3.2-1B & 1.193 / \textbf{1.040} & 1.865 / \textbf{1.590} & 1.197 / \textbf{1.095} & 1.574 / \textbf{1.531} & 1.057 / \textbf{0.908} & 1.420 / \textbf{1.149} & 1.192 / \textbf{1.030} & 1.800 / \textbf{1.667} \\
    Llama-3.1-8B & 0.986 / \textbf{0.938} & 1.641 / \textbf{1.252} & 1.081 / \textbf{1.000} & 1.563 / \textbf{1.373} & 0.862 / \textbf{0.805} & 1.232 / \textbf{0.834} & 1.002 / \textbf{0.883} & 1.714 / \textbf{1.462} \\
    \midrule
    Qwen3-1.7B & 0.906 / \textbf{0.861} & 1.155 / \textbf{1.028} & 0.994 / \textbf{0.871} & 1.128 / \textbf{1.082} & 0.917 / \textbf{0.853} & 1.028 / \textbf{0.868} & 1.040 / \textbf{0.869} & 1.330 / \textbf{1.222} \\
    Qwen3-8B & 0.854 / \textbf{0.792} & 1.058 / \textbf{0.910} & 0.856 / \textbf{0.795} & 1.042 / \textbf{0.982} & 0.831 / \textbf{0.765} & 0.873 / \textbf{0.729} & 0.869 / \textbf{0.772} & 1.184 / \textbf{1.107} \\
    \bottomrule
    \end{tabular}%
    }
    \caption{Performance across model families and sizes on the participant-held-out split. Each cell reports Centaur~/~Cross-Centaur negative log likelihood for the corresponding task family; lower values are better, and the better of the two is in bold.}
    \label{tab:model_variants}
\end{table*}

\section{Prompt Details}

\subsection{Prompt Template}
\label{app:prompt_template}

Cross-Centaur examples are formatted as chat-style supervised fine-tuning instances.
The user message provides the source task description, the source behavior transcript, and the target task description.
The assistant message contains the target behavior transcript to be predicted.

\begin{figure*}[t]
    \centering
    \begin{minipage}{0.95\textwidth}
    \begingroup
    \footnotesize
    \begin{tcolorbox}[colframe=black!70, colback=blue!3!white, coltitle=black, arc=3pt, boxrule=0.4pt, title=User message, left=0pt, right=0pt, top=2pt, bottom=2pt]
Your task is to predict a participant's behavior in a target experiment from their behavior in an input experiment. First read the input experiment description and behavior, then read the target experiment description.\\

\textcolor{teal}{--- PREVIOUS EXPERIMENT DESCRIPTION ---}\\
\textcolor{gray}{\{}source\_task\_description\textcolor{gray}{\}}\\
\textcolor{teal}{--- END PREVIOUS EXPERIMENT DESCRIPTION ---}\\
\textcolor{teal}{--- PREVIOUS EXPERIMENT ---}\\
\textcolor{gray}{\{}source\_behavior\_transcript\textcolor{gray}{\}}\\
\textcolor{teal}{--- END PREVIOUS EXPERIMENT ---}\\
\textcolor{teal}{--- NEXT EXPERIMENT DESCRIPTION ---}\\
\textcolor{gray}{\{}target\_task\_description\textcolor{gray}{\}}\\
\textcolor{teal}{--- END NEXT EXPERIMENT DESCRIPTION ---}\\
Output only the next experiment's transcript lines in the same format.
    \end{tcolorbox}
    \vspace{0.2em}
    \begin{tcolorbox}[colframe=black!70, colback=yellow!6!white, coltitle=black, arc=3pt, boxrule=0.4pt, title=Assistant message, left=0pt, right=0pt, top=2pt, bottom=2pt]
\textcolor{blue}{\{}target\_behavior\_transcript\textcolor{blue}{\}}
    \end{tcolorbox}
    \endgroup
    \end{minipage}
    \caption{Prompt template for Cross-Centaur: The formatting for user and assistant messages during supervised fine-tuning, with color highlighting for key sections.}
    \label{fig:prompt_fullwidth}
\end{figure*}

\subsection{Task Descriptions}
\label{app:prompt_task_descriptions}

The task-description placeholders in Fig.~\ref{fig:prompt_fullwidth} are instantiated with the following descriptions.
The same text is used for both source and target task descriptions, depending on the task assigned to each prompt.

\begingroup
\small
\setlength{\parindent}{0pt}
\setlength{\parskip}{0.35em}

\begin{tcolorbox}[colframe=black!45, colback=gray!3!white, coltitle=black, arc=2pt, boxrule=0.35pt, title=Two-armed bandit, left=4pt, right=4pt, top=3pt, bottom=3pt]
In this task, you will play a kind of gambling game with two options.

One option is a 'safe' option, which always yields zero points.
The other option is a 'risky' option, which can yield varying rewards (positive or negative) each time you choose it.

Your goal is to maximize your total reward by choosing the options wisely.

In this task, you will make 10 choices.

Note that the reward distributions for the options do not change during the task.
So, the characteristics of the safe and risky options remain the same throughout the task.

When reporting choices, output only the selected option type inside the marked answer, such as <<safe>> or <<risky>>.
\end{tcolorbox}

\begin{tcolorbox}[colframe=black!45, colback=gray!3!white, coltitle=black, arc=2pt, boxrule=0.35pt, title=Restless bandit, left=4pt, right=4pt, top=3pt, bottom=3pt]
In this task, you will play a gambling-style game with four options, labeled A, B, C, and D.

Each time you choose an option, you will receive a reward between 0 and 100 points.

Your goal is to maximize your total reward by choosing wisely among the options.

You will make 50 choices in this task.

Importantly, the quality of each option may change over time.
In other words, an option that gives high rewards now may not do so later in the task.

Try to keep track of your experiences and adjust your choices accordingly.

When reporting choices, output only the selected option letter inside the marked answer, such as <<B>>.
\end{tcolorbox}

\begin{tcolorbox}[colframe=black!45, colback=gray!3!white, coltitle=black, arc=2pt, boxrule=0.35pt, title=Correlated bandit, left=4pt, right=4pt, top=3pt, bottom=3pt]
In this task, you will play a kind of gambling game with 30 options.

Each time you choose an option, you will receive a reward.

Your goal is to maximize your total reward by choosing the options wisely.

The 30 options are arranged in a row, and the rewards of neighboring options are correlated.
In other words, options that are close to each other tend to give similar rewards.

Try to use this information to guide your choices and maximize your total reward.

Some sessions are long (10 choices) and some are short (5 choices).

Note that the reward distribution and its range change between sessions.
So, try to learn which options are better in each session and adjust your choices accordingly.

When reporting choices, output only the selected option number inside the marked answer, such as <<8>>.
\end{tcolorbox}

\begin{tcolorbox}[colframe=black!45, colback=gray!3!white, coltitle=black, arc=2pt, boxrule=0.35pt, title=One-shot gamble, left=4pt, right=4pt, top=3pt, bottom=3pt]
In this task, you will be presented with two lottery options and asked to choose one.

Each option has a different probability and reward amount.

Your goal is to maximize your total reward by choosing the better option.

You will make 10 choices in this task.
On each trial, two options (option A and B) will be presented with their respective probabilities (\%) and reward amounts.

Choose the option you would like to play by clicking on it.

After you make 10 choices, the outcomes of your choices will be revealed.

When reporting choices, output only the selected option letter inside the marked answer, such as <<B>>.
\end{tcolorbox}

\begin{tcolorbox}[colframe=black!45, colback=gray!3!white, coltitle=black, arc=2pt, boxrule=0.35pt, title=Temporal discounting, left=4pt, right=4pt, top=3pt, bottom=3pt]
In this task, you will repeatedly choose how much and when to receive or pay money.

This task asks you about your preferences, so there are no right or wrong answers.

Just choose the option that you prefer.

In each question, you will see two options:

1. Immediate Option: Receive/pay a smaller amount of money immediately (or sooner)

2. Delayed Option: Receive/pay a larger amount of money after a delay (or later)

Please choose the option you prefer.

When reporting choices, output only the selected timing label inside the marked answer, such as <<immediate>> or <<delayed>>.
\end{tcolorbox}

\begin{tcolorbox}[colframe=black!45, colback=gray!3!white, coltitle=black, arc=2pt, boxrule=0.35pt, title=Shuffled card task, left=4pt, right=4pt, top=3pt, bottom=3pt]
In this task, you will decide whether to keep drawing cards or stop and keep your current points.

The task begins with 32 facedown cards.

A gain card adds a small number of points.
A loss card subtracts a large number of points and immediately ends the session.

You may stop at any time and keep the points collected so far.

At the start of the task, you will be told the gain amount, the loss amount, and how many loss cards are in the deck.

When reporting turned cards, output only the selected card number inside the marked answer, such as <<8>>.
\end{tcolorbox}

\begin{tcolorbox}[colframe=black!45, colback=gray!3!white, coltitle=black, arc=2pt, boxrule=0.35pt, title=Classification, left=4pt, right=4pt, top=3pt, bottom=3pt]
In this task, you will see a series of figures and classify each into one of two groups.

Each figure differs by shape (triangle or square), color (red or green), and size (small or large).

The figures differ in shape (triangle or square), color (red or green), and size (small or large).

There are 8 possible figures, and their group (A or B) is determined by a combination of these features. Try to figure out the rule as you go.

You will complete 16 classification trials.

You will do 16 classification trials. Try to classify as accurately as possible by inferring the rule.

When reporting choices, output only the selected group letter inside the marked answer, such as <<A>> or <<B>>.
\end{tcolorbox}

\begin{tcolorbox}[colframe=black!45, colback=gray!3!white, coltitle=black, arc=2pt, boxrule=0.35pt, title=Estimation, left=4pt, right=4pt, top=3pt, bottom=3pt]
In this task, you will predict one medical test result using two numerical results from two other tests.

On each trial, you will be shown two cue values and asked to estimate the target value.

Your goal is to make your estimates as accurate as possible.

This task uses medical tests for three fictitious hormones:
Progladine, Amalydine, and Caldionine.

You will see two numerical results from two tests: one measuring Progladine and the other measuring Amalydine.
These two values are known to be associated with the level of Caldionine, which you will estimate in this task.

The Caldionine value you need to estimate ranges from 10 to 90, in increments of 10.

The Progladine and Amalydine values will vary from 1 to 5 in each trial.

You will experience 25 different patterns of Progladine and Amalydine values.
After submitting your estimate, you will receive the true Caldionine value as feedback.

When reporting estimates, output only the estimated Caldionine value inside the marked answer, such as <<40>>.
\end{tcolorbox}

\begin{tcolorbox}[colframe=black!45, colback=gray!3!white, coltitle=black, arc=2pt, boxrule=0.35pt, title=Odd one out, left=4pt, right=4pt, top=3pt, bottom=3pt]
In this task, you will be shown three pictures and asked to identify the one that is different from the other two.

This task asks you about your thinking style, so there are no right or wrong answers.

In each trial, you will see three pictures.
These pictures include various kinds of objects such as animals, food, vehicles, and so on.

You will make 20 choices in this task.

When reporting choices, output only the selected picture name inside the marked answer, such as <<pothole>>.
\end{tcolorbox}

\begin{tcolorbox}[colframe=black!45, colback=gray!3!white, coltitle=black, arc=2pt, boxrule=0.35pt, title=Public goods game, left=4pt, right=4pt, top=3pt, bottom=3pt]
In this task, you will decide how many tokens to contribute to a group project with three other participants on Prolific.

Your goal is to maximize your total earnings by choosing how many tokens to put into the group project each round.

Each round, you receive 20 tokens and can contribute any whole-number amount from 0 to 20.

The total group contribution is multiplied by 1.5 and divided equally among the four participants (including you).

For example, if you put in 10 tokens and the others put in a total of 30, the group pool is 40 $\times$ 1.5 = 60 tokens, and each person gets 15 (60 $\div$ 4). Your total for the round is 20 - 10 + 15 = 25 tokens.

If you put in 0 and the others put in 20, the group pool is 20 $\times$ 1.5 = 30 tokens, so each person gets 7.5 (30 $\div$ 4). Your total for the round is 20 - 0 + 7.5 = 27.5 tokens.

This task has two parts:

- In the first part, you choose how many tokens to contribute without seeing what the others do.
- In the second part, you will see the average contribution of the other three participants and decide how much you would contribute at each average value.

Your final earnings will be based on your choices and those of three randomly matched participants.

When reporting contributions, output only the selected token amount inside the marked answer, such as <<5>>.
\end{tcolorbox}

\begin{tcolorbox}[colframe=black!45, colback=gray!3!white, coltitle=black, arc=2pt, boxrule=0.35pt, title=Social value orientation, left=4pt, right=4pt, top=3pt, bottom=3pt]
In this task, you will make a series of decisions about allocating money between yourself and another person.

Imagine that you have been randomly paired with the other person.
This other person is someone you do not know, and you will remain anonymous to each other.

You will be presented with 15 allocation decisions.

In each decision, you will see 9 different ways to allocate money between you and the other person.
Indicate the allocation you prefer most by moving the slider.

This task asks you about your preferences for allocating money.

Just choose the allocation that you prefer.

When reporting choices, output only the selected allocation number inside the marked answer, such as <<3>>.
\end{tcolorbox}

\begin{tcolorbox}[colframe=black!45, colback=gray!3!white, coltitle=black, arc=2pt, boxrule=0.35pt, title=Ultimatum game, left=4pt, right=4pt, top=3pt, bottom=3pt]
In this task, you will allocate points between yourself and another participant.

Imagine you have 100 points to split between you and another participant.
You will play as both the "proposer" and the "responder".

As the proposer, you can choose how to split the points between yourself and the responder.
In this role, you will indicate how many points you want to give to the responder.

If your offer is accepted by the responder, the points will be split according to your offer. However, if the offer is rejected, neither you nor the responder will earn any points.

Conversely, you will also play as the responder.
In this role, you will indicate the minimum offer you are willing to accept from the proposer.

If you accept the offer from the proposer, the points will be split according to the proposer's offer. If you reject the offer, neither you nor the proposer will get any points.

Let's look at an example.

Imagine you are the proposer and you offer 30 points to the responder.
If the responder's acceptance threshold is 20 points, your offer will be accepted, and you will earn 70 points (while the responder earns 30 points).

However, if the responder's acceptance threshold is 40 points, your offer will be rejected, and you will earn zero points (while the responder also earns zero points).

Let's see another example.

Imagine you are the responder and your acceptance threshold is 60 points.
If the proposer offers you 60 points, you will accept the offer and earn 60 points (while the proposer earns 40 points).

However, if the proposer offers you 55 points, you will reject the offer and earn zero points (while the proposer also earns zero points).

After completing this task, you will be randomly paired with another participant and evaluated in both roles (proposer and responder).
As the proposer, your offer will be compared to your partner's acceptance threshold. As the responder, your acceptance threshold will be compared to the offer made by your partner.

When reporting offers or acceptance thresholds, output only the selected point amount inside the marked answer, such as <<50>>.
\end{tcolorbox}

\endgroup

\section{AI Assistant Usage}
We used an AI assistant to check for grammatical errors and language clarity throughout the manuscript drafts.

\label{sec:appendix}

\end{document}